%% file: main.tex
\documentclass{article}

\usepackage{microtype}
\usepackage{graphicx}
\usepackage{subfigure}
\usepackage{booktabs} % for professional tables

\usepackage{hyperref}

\usepackage[accepted]{icml2025}

\usepackage{amsmath}
\usepackage{amssymb}
\usepackage{mathtools}
\usepackage{amsthm}

\usepackage[capitalize,noabbrev]{cleveref}

\theoremstyle{plain}

\theoremstyle{definition}

\theoremstyle{remark}

\usepackage[textsize=tiny]{todonotes}

\icmltitlerunning{Submission and Formatting Instructions for ICML 2025}
\usepackage{graphicx}
\usepackage{makecell}
\usepackage{multirow}
\usepackage{amssymb}
\usepackage{marvosym}
\usepackage{hyperref}
\usepackage{amsmath}
\usepackage{url}
\usepackage{booktabs}       % professional-quality tables
\usepackage{amsfonts}       % blackboard math symbols
\usepackage{nicefrac}       % compact symbols for 1/2, etc.
\usepackage{microtype}      % microtypography
\usepackage{xcolor}         % colors
\usepackage{adjustbox}
\usepackage{float}
\usepackage{graphicx}
\usepackage{comment}
\usepackage{amsmath,amssymb} %
\usepackage{color, soul}
\usepackage{colortbl}
\usepackage{enumitem}
\usepackage{tcolorbox}
\usepackage{subfigure}
\usepackage{graphicx}  % 用于插入图片
\usepackage{subcaption}  % 用于创建子图
\usepackage{xspace}
\usepackage{amsthm}
\usepackage{multirow}
\usepackage{arydshln} %表格虚线
\usepackage{textcomp}
\usepackage{mathtools}
\usepackage{bbm}
\usepackage{wrapfig}
\usepackage{makecell, verbatim, sidecap}
\usepackage{multirow}
\usepackage{nicefrac}
\usepackage{gensymb}
\usepackage{algorithm}
\usepackage{algorithmic}
\usepackage{bm}

\usepackage[T1]{fontenc}
\usepackage[utf8]{inputenc}

\usepackage{microtype}

\usepackage{inconsolata}
\usepackage{pifont}% http://ctan.org/pkg/pifont
\usepackage{enumitem}
\let\oldding\ding% Store old \ding in \oldding
\renewcommand{\ding}[2][1]{\scalebox{#1}{\oldding{#2}}}

\begin{document}

\twocolumn[
% \icmltitle{Submission and Formatting Instructions for \\
%           International Conference on Machine Learning (ICML 2025)}
\icmltitle{RWKVQuant: Quantizing the RWKV Family with Proxy Guided Hybrid of Scalar and Vector Quantization}

% It is OKAY to include author information, even for blind
% submissions: the style file will automatically remove it for you
% unless you've provided the [accepted] option to the icml2025
% package.

% List of affiliations: The first argument should be a (short)
% identifier you will use later to specify author affiliations
% Academic affiliations should list Department, University, City, Region, Country
% Industry affiliations should list Company, City, Region, Country

% You can specify symbols, otherwise they are numbered in order.
% Ideally, you should not use this facility. Affiliations will be numbered
% in order of appearance and this is the preferred way.
\icmlsetsymbol{equal}{*}

\begin{icmlauthorlist}
\icmlauthor{Chen Xu}{equal,houmo}
\icmlauthor{Yuxuan Yue}{equal,houmo,hitsz}
\icmlauthor{Zukang Xu}{houmo}
\icmlauthor{Xing Hu}{houmo}\\
\icmlauthor{Jiangyong Yu}{houmo}
\icmlauthor{Zhixuan Chen}{houmo}
\icmlauthor{Sifan Zhou}{houmo}
%\icmlauthor{}{sch}
\icmlauthor{Zhihang Yuan}{houmo}
\icmlauthor{Dawei Yang$^{\textrm{\Letter}}$}{houmo}
%\icmlauthor{}{sch}
%\icmlauthor{}{sch}
\end{icmlauthorlist}

\icmlaffiliation{houmo}{Houmo AI}
\icmlaffiliation{hitsz}{Harbin Institute of Technology (Shenzhen)}
% \icmlaffiliation{intern}{This work was conducted during his internship at Houmo AI}

\icmlcorrespondingauthor{Dawei Yang}{dawei.yang@houmo.ai}
% \icmlcorrespondingauthor{Firstname2 Lastname2}{first2.last2@www.uk}

% You may provide any keywords that you
% find helpful for describing your paper; these are used to populate
% the "keywords" metadata in the PDF but will not be shown in the document
\icmlkeywords{Machine Learning, ICML}
\vskip 0.3in
]

% this must go after the closing bracket ] following \twocolumn[ ...

% This command actually creates the footnote in the first column
% listing the affiliations and the copyright notice.
% The command takes one argument, which is text to display at the start of the footnote.
% The \icmlEqualContribution command is standard text for equal contribution.
% Remove it (just {}) if you do not need this facility.

%\printAffiliationsAndNotice{}  % leave blank if no need to mention equal contribution
\printAffiliationsAndNotice{\icmlEqualContribution} % otherwise use the standard text.

\input{texs/0_Abstract}
\input{texs/1_intro_refine}
\input{texs/3_preliminaries}
\input{texs/4_method}
\input{texs/5_experiments}

\input{texs/6_conclusion}

\section*{Impact Statement}
This paper aims to promote the application of the RWKV family, mainly focused on the post training quantization methods. By introducing RWKVQuant, our approach enables the deployment of RWKV models on resource-constrained devices.

% In the unusual situation where you want a paper to appear in the
% references without citing it in the main text, use \nocite
\nocite{langley00}

\bibliography{example_paper}
\bibliographystyle{icml2025}

%%%%%%%%%%%%%%%%%%%%%%%%%%%%%%%%%%%%%%%%%%%%%%%%%%%%%%%%%%%%%%%%%%%%%%%%%%%%%%%
%%%%%%%%%%%%%%%%%%%%%%%%%%%%%%%%%%%%%%%%%%%%%%%%%%%%%%%%%%%%%%%%%%%%%%%%%%%%%%%
\input{texs/7_Appendix}

%%%%%%%%%%%%%%%%%%%%%%%%%%%%%%%%%%%%%%%%%%%%%%%%%%%%%%%%%%%%%%%%%%%%%%%%%%%%%%%
%%%%%%%%%%%%%%%%%%%%%%%%%%%%%%%%%%%%%%%%%%%%%%%%%%%%%%%%%%%%%%%%%%%%%%%%%%%%%%%
%\newpage
%\appendix
%\onecolumn
%\section{APPENDIX}
%\input{texs/7_Appendix}

%You can have as much text here as you want. The main body must be at most $8$ pages long.
%For the final version, one more page can be added.
%If you want, you can use an appendix like this one.  

%The $\mathtt{\backslash onecolumn}$ command above can be kept in place if you prefer a one-column appendix, or can be removed if you prefer a two-column appendix.  Apart from this possible change, the style (font size, spacing, margins, page numbering, etc.) should be kept the same as the main body.
%%%%%%%%%%%%%%%%%%%%%%%%%%%%%%%%%%%%%%%%%%%%%%%%%%%%%%%%%%%%%%%%%%%%%%%%%%%%%%%
%%%%%%%%%%%%%%%%%%%%%%%%%%%%%%%%%%%%%%%%%%%%%%%%%%%%%%%%%%%%%%%%%%%%%%%%%%%%%%%

\end{document}

%% file: texs/0_Abstract.tex
\begin{abstract}\label{sec_abstract}
RWKV is a modern RNN architecture with comparable performance to Transformer, but still faces challenges when deployed to resource-constrained devices. 
Post Training Quantization (PTQ), which is a an essential technique to reduce model size and inference latency, has been widely used in Transformer models.
However, it suffers significant degradation of performance when applied to RWKV.
This paper investigates and identifies two key constraints inherent in the properties of RWKV:  (1) Non-linear operators hinder the parameter-fusion of both smooth- and rotation-based quantization, introducing extra computation overhead. (2) The larger amount of uniformly distributed weights poses challenges for cluster-based quantization, leading to reduced accuracy.
To this end, we propose RWKVQuant, a PTQ framework tailored for RWKV models, consisting of two novel techniques: (1) a coarse-to-fine proxy capable of adaptively selecting different quantization approaches by assessing the uniformity and identifying outliers in the weights, and (2) a codebook optimization algorithm that enhances the performance of cluster-based quantization methods for element-wise multiplication in RWKV.
Experiments show that RWKVQuant can quantize RWKV-6-14B into about 3-bit with less than 1\% accuracy loss and 2.14$\times$ speed up.
\end{abstract}

%% file: texs/1_intro_refine.tex
\section{Introduction}\label{sec_intro}
RWKV~\cite{peng2023rwkv} is a modern sequence model that integrates the strengths of both Recurrent Neural Networks (RNNs)~\cite{elman1990finding} and Transformer~\cite{vaswani2017attention}.
It has a comparable capacity to Transformer-based Large Language Models (T-LLMs) while retaining the efficient inference feature of RNNs, positioning it a promising foundational architecture for both language~\cite{peng2024eagle} and vision~\cite{zhou2024bsbp} tasks. 
\begin{figure}[!h]
    \centering
    \includegraphics[width=0.4\textwidth]{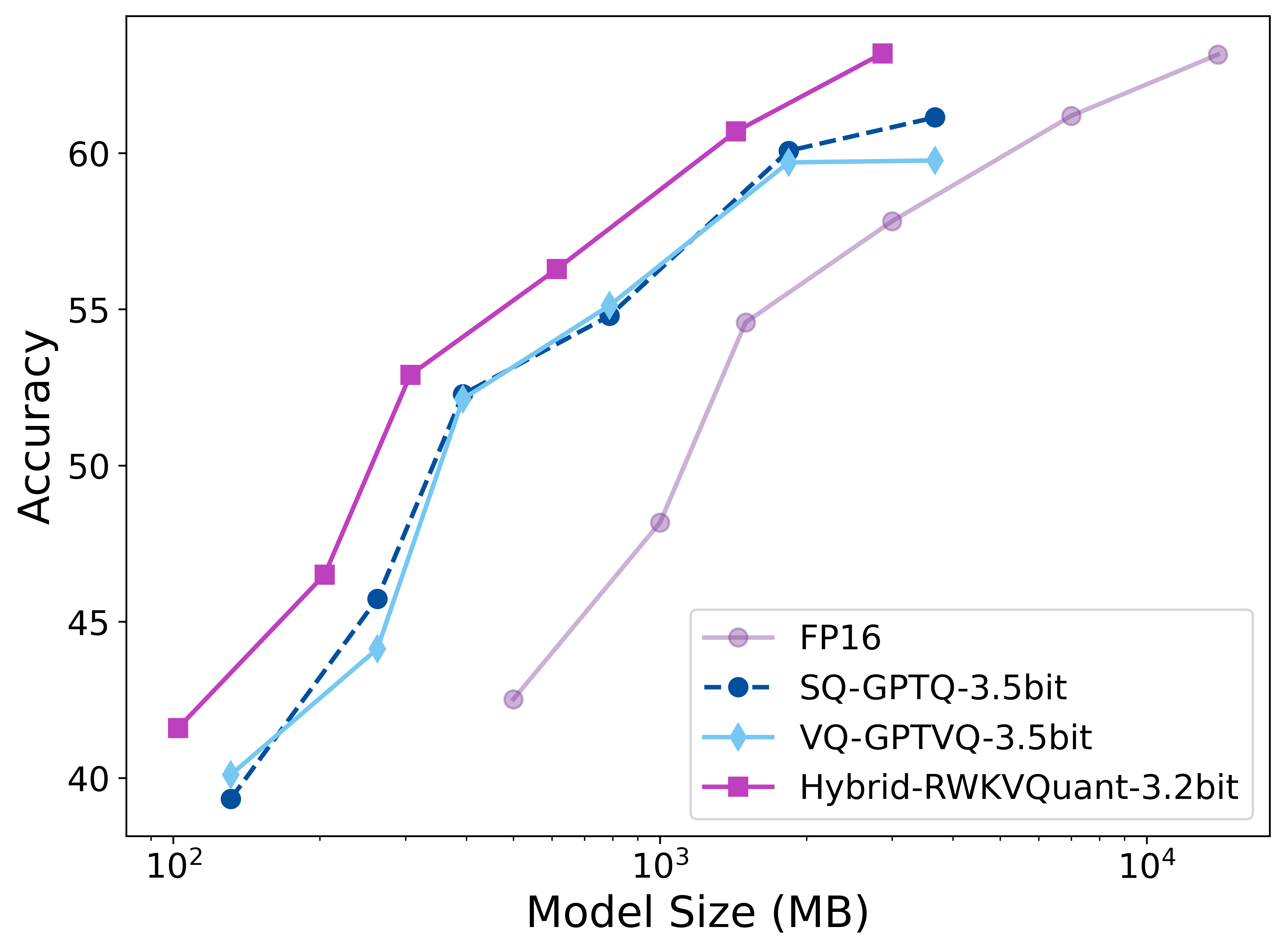}
    \caption{Accuracy-model size curve. Results of zero-shot accuracy are evaluated on the LAMBADA dataset~\cite{radford2019language}. 
    Our proposed RWKVQuant outperforms the individual utilization of SQ and VQ methods for all sizes of models. 
    }
    \vspace{-2mm}
    \label{fig_performance_compare}
\end{figure}

Despite the advantages, its vast size of parameters have posed a significant barrier to the deployment on resource-constrained devices. 
For instance, the compute-to-memory access ratio (FLOPs/Bytes) of RWKV-6-7B~\cite{peng2024eagle} is 0.97, while that for the decoding phase of LLaMA-2-7B~\cite{touvron2023llama2} is 4.88 (detailed in \ref{appendix a}). 
Secondly, large RWKV models demand substantial memory resources.  
For instance, RWKV-6-14B requires approximately 30GB of memory to be loaded, which typically exceeds the capacity of edge devices. 

Post Training Quantization (PTQ), including Scaler Quanzization (SQ) and Vector Quantization (VQ), is a widely adopted approach to reduce model size and inference latency for T-LLMs~\cite{2023omniquant, ashkboos2024quarot, ostquant, yuan2024llm}.
However, directly applying the most advanced quantization frameworks to RWKV models leads to severe performance degradation.
For instance, applying QuaRot~\cite{ashkboos2024quarot} (belongs to SQ) to RWKV-7 series models increases the overall FLOP by more than 99\%, and applying VPTQ~\cite{liu2024vptq} (belongs to VQ) to RWKV-6-7B model leads to more than 16\% accuracy decline. 

In depth, we investigate and identify two primary limitations inherent in the properties of RWKV. 
\raisebox{-0.5pt}{\ding[1.1]{182\relax}} 
\textbf{Non-linear operators hinder the parameter-fusion of both smooth- and rotation-based methods. }
Typically, these SQ approaches fuse the introduced parameters, i.e., smoothing vectors and orthogonal matrices, into neighbored normalization layers and linear layers of T-LLMs. 
However, the RWKV structure employs several non-linear operators along the fusion path, including token-shift, Sigmoid function, and exponential function. 
These modules can block the linear fusion process, inevitably leading to additional runtime overhead.
\raisebox{-0.5pt}{\ding[1.1]{183\relax}} 
\textbf{The larger amount of uniformly distributed weights poses challenges for cluster-based quantization. } 
While such VQ methods benefit from distinctly categorized distribution, RWKV tends to have more uniform weights compared to T-LLMs (detailed in Section~\ref{subsec_analysis_rwkv_uniformity}), which complicates the clustering process as shown in Table~\ref{tab_kmeans_loss}. 
\begin{table}[!t]
\caption{The average relative cluster loss of weights for the RWKV and LLaMA family, computed by KMeans~\cite{lloyd1982least}. }
\vskip 0.15in
\centering
\begin{tabular}{cccc}
\toprule
Family & Model & 8 Clusters & 16 Clusters \\ \midrule
\multirow{2}{*}{RWKV} & 6-7B & 2.01 & 0.78 \\
     & 6-14B & 1.98 & 0.78 \\ \midrule
\multirow{2}{*}{LLaMA} & 2-7B & 0.96 & 0.65 \\
     & 2-14B & 0.89 & 0.64 \\ \bottomrule
\end{tabular}
\label{tab_kmeans_loss}
\end{table}

To this end, we propose RWKVQuant, an effective and efficient post-training quantization (PTQ) framework tailored for RWKV models. 
Our core insight is to enhance VQ by partially applying the classic compensation-based SQ methods like GPTQ~\cite{frantar2022gptq}, which are more suitable for uniformly distributed weights. 
Specifically, we propose a coarse-to-fine proxy to optimize the hybrid strategy. 
(1) The coarse-grained proxy is established on the basis of Information Entropy (IE)~\cite{shannon1948mathematical}, which evaluate the overall uniformity. For non-uniform weights, VQ is directly applied. 
(2) For uniform weights, we further introduce a fine-grained proxy, computed by weighted hight-order central moments, to detect local outliers. VQ is applied when outliers emerge; otherwise, SQ is applied. 
In addition to the hybrid, we further optimize VQ for the unique element-wise multiplication operator of RWKV. 

Experiments show that RWKVQuant outperforms existing methods across various tasks on different
RWKV model families, including RWKV-6~\cite{peng2024eagle} and RWKV-7~\cite{peng_bo_2021_5196578} for
RWKV-based language tasks, as well as VRWKV~\cite{duan2024vision} for RWKV-based vision tasks. 
As shown in Figure~\ref{fig_performance_compare}, RWKVQuant quantizes weights into about 3-bit and achieves superior accuracy compared to the individual utilization of SQ and VQ. 
Additionally, RWKVQuant demonstrates remarkable efficiency. 
For instance, it can quantize RWKV-6-14B with less than 1\% accuracy loss, 2.83$\times$ memory saving, and 2.14$\times$ speed up. 
Lastly, our contributions can be concluded as follows. 
\begin{itemize}
    \item We reveal that both smooth- and rotation-based PTQ methods are not well-suitable for RWKV, primarily due to the unavoidable runtime overhead. 
    Further, cluster-based PTQ methods suffer severe accuracy drop, owing to the larger amount of uniformly distributed weights. 
    \item We propose RWKVQuant, which enhance VQ by partially adopting compensation-based SQ methods. 
    It introduces a coarse-to-fine proxy to guide the hybrid strategy. 
    It further enhances the VQ for the unique element-wise multiplication modules in RWKV.
    \item RWKVQuant can effectively and efficiently quantize weights into about 3-bit and outperforms both SQ and VQ methods as shown in Figure~\ref{fig_performance_compare}. 
    \item To the best of our knowledge, RWKVQuant is the first comprehensive PTQ framework for the RWKV family. 
    As a pioneering study, we will publish the \textcolor{red}{\href{https://anonymous.4open.science/r/RWKVQuant-5B27/README.md}{code}} in the hope of promoting further research and facilitating advancements in this field.
\end{itemize}

%% file: texs/3_preliminaries.tex
\section{Preliminaries}\label{sec_preliminaries}
\subsection{RWKV Structure and Models}\label{subsec_rwkv_structure}
\begin{figure}[!t]
    \centering
    %这张图换成矢量图，目前来看一放大分辨率就有些糊。
    \includegraphics[width=0.47\textwidth]{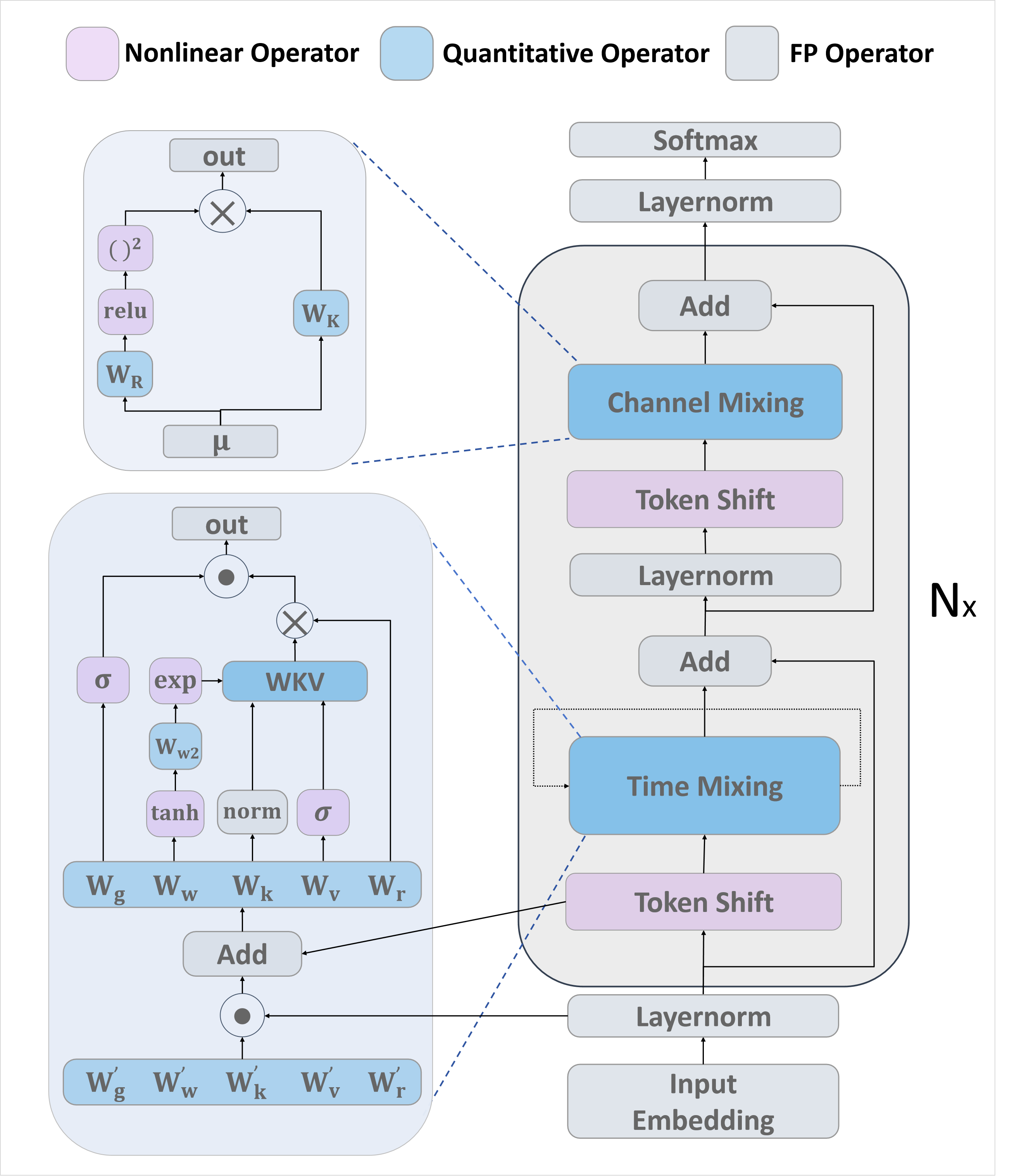}
    \caption{Model Structure of RWKV-7. It contains several blocks and each block has two key modules: Time Mixing and Channel Mixing. }
    \label{fig_rwkv_struture}
\end{figure}

Referring to Figure~\ref{fig_rwkv_struture}, RWKV structure contains two key modules, including Time Mixing and Channel Mixing (detailed in \ref{appendix_rwkv_structure}). 
With the previous word $\bm{x}_{t-1}$, the current word $\bm{x}_t$ and can be derived by a token-shift operator:
\begin{equation}
    \bm{x}_t=\mathrm{concat}(\bm{x}_{t-1}[1:,], \mathbf{0}),
    \label{eq_token_shift}
\end{equation}
where $\mathbf{0}$ denotes an all-zero vector. 
RWKV models make use of Time Mixing for seizing the relationship among tokens and utilize Channel Mixing to probe the dimensions within the hidden layer that are relevant to individual tokens. 

Compared to T-LLMs, these modifications enables RWKV to decrease substantially the computational overhead and memory demands while effectively retaining the capacity to model long-term dependencies. 
Thereby, the RWKV family has already manifested its potential in a diverse array of real-world applications~\cite{li2024survey}, including QQ~\cite{qq}, WeChat~\cite{wechat1, wechat2}, and Telegram~\cite{telegram}. 
% nlu
For natural language understanding tasks, RWKV-v6~\cite{peng2024eagle} has achieved remarkable advancements in accuracy, attaining comparable performance to those of larger models like SomlLM and Qwen~\cite{chen2024onlysportslm}.
% vision
For vision tasks, BSBP-RWKV~\cite{zhou2024bsbp} excels in the domain of image segmentation. 
% % It integrates conventional noise reduction methods with a resilient network architecture, thereby attaining superior accuracy and performance. 
% % In a similar vein, Restore-RWKV~\cite{yang2024restore} stands out in the restoration of medical images. It incorporates a novel all-directional token translation layer and a reconstructive attention mechanism, effectively capturing the spatial correlations within 2D imagery. %这里的Restore-RWKV可以删去，不仅没中稿还是医疗图像任务的，和我们的任务没什么关系。不用花这么多篇幅叙述他。
% % 缺点
\subsection{Post Training Quantization (PTQ)}\label{subsec_ptq_works}
PTQ serves as a potent strategy for model compression. 
By converting the high-precision variables of pre-trained models into low-bit integers, it achieves a reduction in memory usage and an acceleration of inference speed.
Typically, PTQ can be divided into two main approaches: SQ and VQ. 
% SQ is generally performed to obtain a low-precision representation (e.g., 4-bit integer) from a high-precision variable (e.g., 16-bit floating points). 
\paragraph{Scalar Quantization (SQ)} 
SQ maps the original data to the quantized range by a scaling factor, and subsequently rounds floating points to the uniform-distributed integers. For a tensor $\bm{x}$ to be quantized, it can be uniformly quantized to $b$-bits as follows~\citep{Jacob_2018_CVPR}:
\begin{equation}
    Q(\bm{x})=\mathrm{clamp}(\lfloor \frac{\bm{x}}{s} \rceil+{z}, 0, 2^b-1),
    \label{eq_SQ}
\end{equation}
where $Q(\cdot)$ represents the quantization function, $s=(\max(\bm{x})-\min(\bm{x}))/2^b-1$ is the scale factor,  ${z}=-\min(\bm{x})/s$ is the zero point, $\lfloor \cdot \rceil$ denotes the rounding-to-nearest operator, and $\mathrm{clamp}$ is the clipping function. 

SQ is widely-adopted by most of the PTQ frameworks \cite{yang2024post}. For instance, the classic compensation-based GPTQ~\cite{frantar2022gptq} can quantize weights to 3-4 bit with slight accuracy drop based on approximate second-order information. 
To address outliers, AWQ~\cite{lin2023awq}, SmoothQuant~\cite{xiao2022smoothquant}, and OmniQuant~\cite{2023omniquant} explore the scheme of smoothing by detecting the importance of different activation channels. 
Recent works (e.g., Quarot~\cite{ashkboos2024quarot}, SpinQuant~\cite{liu2024spinquant}, and OSTQuant~\cite{ostquant}) further suppress outliers by rotating the variables to be quantized with orthogonal matrices. 
% Both smoothing and rotation are intended to make the original distribution more uniform, thus reducing the rounding error of SQ.

% \label{subsec_vector_quant}
\paragraph{Vector Quantization (VQ)} 
VQ quantizes several vectors into a finite subset, which is commonly referred to as a codebook $\bm{C}$ ~\cite{gersho1979asymptotically}. Typically it has shape ($2^k, d$), where $k$ is the bits of the index and $d$ is the vector dimension. 
Given a tensor $\bm{x}$ with shape ($m, n$) to be quantized, VQ first transforms it into $x^\prime$ with dimensions ($m * n // d, d$). 
Second, for each $d$-dimensional vector in $\bm{x}^\prime$, VQ replaces it with the $k$-bit index of the nearest vector from the codebook.
For instance, if we use the Euclidean distance (calculated by the Frobenius normalization ${||\cdot||}_F$) to measure similarities, the quantization process can be expressed as:
\begin{equation}
    Q(\bm{x}^\prime)=\{\underset{j \in 2^k}{\mathrm{argmin}} {||\bm{x}_i^\prime - \bm{C}_j||}_F\thinspace|\thinspace i=1,...,m*n//d\}. 
    \label{eq_vq}
\end{equation} 

Compared to SQ, this scheme takes the advantage of maintaining the shape of the source distribution, especially under lower bit-width. 
For example, VPTQ~\cite{liu2024vptq} and GPTVQ~\cite{van2024gptvq} combine VQ with GPTQ, achieving advanced performances under 2$\sim$3 bits. To obtain the codebook, they cluster the source vectors by K-Means Algorithm~\cite{lloyd1982least} and Expectation-Maximization Algorithm~\cite{moon1996EM}, respectively. AQLM~\cite{egiazarian2024extreme} further utilizes layer-wise training for the codebook to obtain optimal accuracy. 
% These methods essentially represent the original data source in the form of clustering.
% It first reshapes the original tensor into vectors, then maps them to a finite subset, which is commonly referred to as a codebook.
% While both SQ and VQ perform effectively for Transformer-based large language models, they do not work well with RWKV models. 
% 要不要说这一句↓ 
% 这一句可以酌情放到Intorduction里.related work里放在此处，一般情况下reviewer看不太到.
% Notably, to our knowledge, our method is the first PTQ solution specifically designed for the RWKV family, applicable to both RWKV-based language and vision tasks.

%% file: texs/4_method.tex
\section{Method}\label{sec_method}
\subsection{Coarse-to-fine Proxy for Hybrid Quantization}\label{subsec_hybrid}
\paragraph{Hybrid of SQ and VQ}
% % 实验观察发现
% From visualization and experiments, we observe that parts of weights in RWKV models exhibit an obviously uniform distribution. Applying SQ methods, such as GPTQ~\cite{frantar2022gptq}, to these parts achieves a superior performance compared to VQ methods as shown in Figure~\ref{fig_uniform_weight}. 
% On the other hand, other parts appear to be generally even, yet with local outliers. Applying VQ methods, such as GPTVQ~\cite{van2024gptvq}, to these weights outperforms SQ methods as shown in Figure~\ref{fig_uneven_weight}.

% % 解释原因
% An intuitive reason is that, SQ directly maps the original data to fixed intervals. A uniform data can equally maximize the representational ability of all quantized integers. When outliers emerge, certain integers become overloaded, whereas others remain underutilized, resulting in significant reduction of the overall efficiency of the system.
% On the contrary, VQ can allocate clusters to outliers, thus maintaining the quantization performance. However, it is challenging for VQ to derive distinctive clusters from a uniform data, which diminishes the representational capacity of the codebook.
% 引出动机
Given inputs $\bm{x}$, weights $\bm{\theta}$, number of weights $M$, and the model $f(\cdot)$, the optimization goal is to minimize the expectation $\mathbb{E}[\cdot]$ of the Mean Square Error (MSE) of the model output:
\begin{equation}
\label{eq_optimization_goal}
    \begin{aligned}
        &\underset{\bm{\phi}}{\mathrm{argmin}}\thinspace \mathbb{E}[\thinspace{||f_{\bm{\theta}}(\bm{x})-f_{\bm{\theta}^\prime}(\bm{x})||}_F^2\thinspace] \\
        &\mathrm{s.t.} \bm{\phi}=\{\phi_m\in\{0,1\}\thinspace|\thinspace m=1,2,...,M\}\\
        &\quad\thinspace\bm{\theta}^\prime=\{\phi_m\mathrm{SQ}(\bm{\theta}_m)+(1-\phi_m)\mathrm{VQ}(\bm{\theta}_m)\thinspace\\
        &\qquad\qquad|\thinspace\bm{\theta}_m\in\bm{\theta}, m=1,2,...,M\}.
    \end{aligned}
\end{equation}
Here, $\bm{\phi}$ represents the collection of options for SQ and VQ. Although the optimal solution of Equation~\ref{eq_optimization_goal} can be found by the exhaustive algorithm, its complexity increases exponentially with the number of weights, i.e., $O(2^M)$. Considering the computational cost, we construct an effective proxy by evaluating the uniformity and outliers of each weight from both coarse- and fine-grained perspectives, whose complexity decreases to $O(M)$.

\paragraph{Coarse-grained Proxy}
% 引入信息熵
Information Entropy (IE)~\cite{shannon1948mathematical} is one of the most common approaches to evaluate uniformity. 
However, it measures the probability distribution, rather than the original data that concerned by quantization. 
To take advantage of its effectiveness, we perform a series of transformations on the model weights.

% 变换过程
Given a weight $\bm{W}\in\mathbb{R}^{oc\times ic}$, it is first flattened, then sorted in ascending order to formulate $\bm{W}^\prime\in\mathbb{R}^{oc\cdot ic}$. 
Subsequently, the intervals $\bm{G}\in\mathbb{R}^{(oc\cdot ic)-1}$ of all adjacent positions in $\bm{W}^\prime$ can be calculated by:
\begin{equation}
    \bm{G} = \bm{W}^\prime[1:]-\bm{W}^\prime[:-1].
    \label{eq_interval}
\end{equation} 
For the clarity of expression, the term $(oc\cdot ic)-1$ is denoted by the symbol $n$ in the following contents. Next, $\bm{G}$ is transformed to $\bm{G}^\prime$ by:
\begin{equation}
    \bm{G}^\prime=\{\bm{G}_i^\prime=\frac{\bm{G}_i}{\sum_{i=1}^{n}\bm{G}_i}\thinspace|\thinspace i=1,2,...,n\}.
    \label{eq_interval_norm}
\end{equation}
Considering that $\bm{G}^\prime$ satisfies that $\sum_{i=1}^{n}\bm{G}_i^\prime=1$, it can be treated as a discrete probability distribution. 
Consequently, its IE (denoted by $H$) can be obtained by:
\begin{equation}
    H(\bm{G}^\prime)=-\sum_{i=1}^{n}\bm{G}_i^\prime\log \bm{G}_i^\prime.
    \label{eq_ie}
\end{equation}
According to the property of IE, Equation~\ref{eq_ie} measures the concentration of $\bm{G}^\prime$.
Since $\bm{G}^\prime$ are the intervals, its concentration can equivalently reflect the uniformity of the original weight $\bm{W}$.  
\begin{figure*}[!t]
    \centering
    \includegraphics[width=0.9\textwidth]{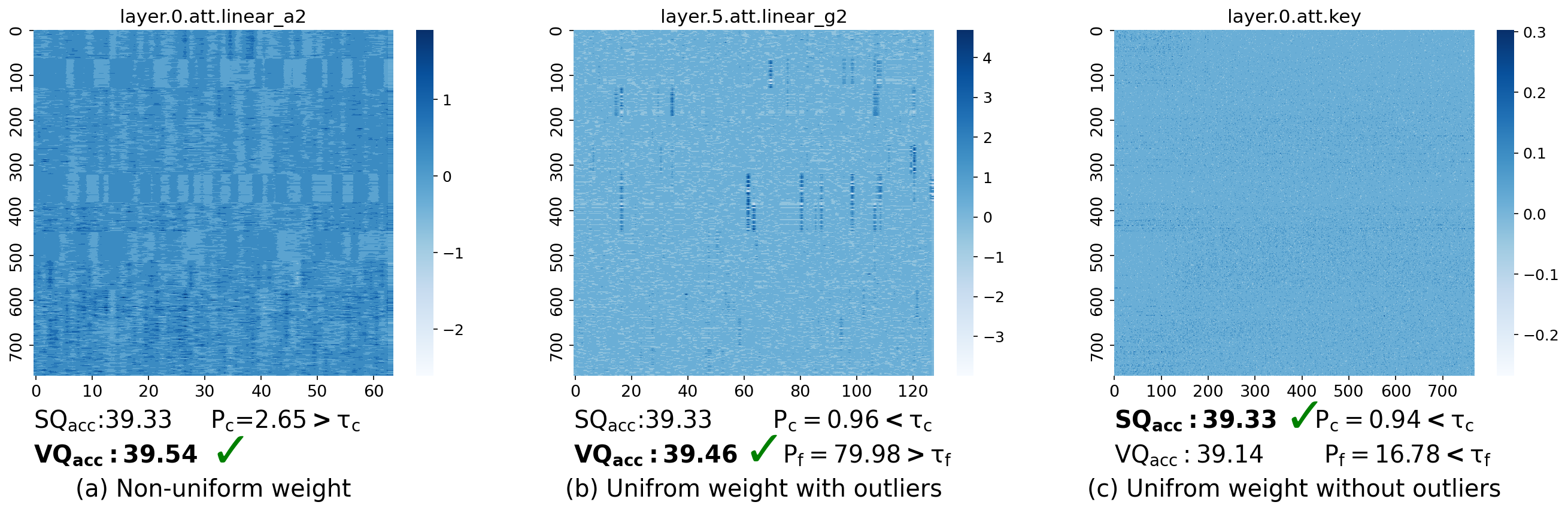}
    \caption{Zero-shot accuracy when applying different quantization methods to specific weights. For the weight in each sub-figure, $\mathrm{SQ}_{\mathrm{acc}}$ denotes the performance when SQ is applied, $\mathrm{VQ}_{\mathrm{acc}}$ denotes the performance when VQ is applied, while all other weights are quantized with VQ. $P_c$ and $P_f$ are coarse- and fine-grained proxy, while $\mu_c$ and $\mu_f$ are their corresponding thresholds. }
    \label{fig_proxy_methods}
\end{figure*}

Assuming an absolutely uniform weight $\hat{\bm{W}}$ with fixed intervals, it can be transformed to $\hat{\bm{G}^\prime}$ following the above process, which finally should be:
\begin{equation}
    \hat{\bm{G}^\prime}=\{\hat{\bm{G}^\prime}_i=\frac{1}{n}\thinspace|\thinspace i=1,2,...,n\}.
    \label{eq_absolute_case}
\end{equation}
Owing to the property of IE, only if $\bm{W}=\hat{\bm{W}}$ dose Equation~\ref{eq_ie} take the maximum value. Finally, the coarse-grained proxy $P_c$ can be obtained by computing the gap between the IE of $\bm{G}^\prime$ and $\hat{\bm{G}^\prime}$:
\begin{equation}
\label{eq_coarse_proxy}
    \begin{aligned}
    P_c(\bm{G}^\prime)=H(\hat{\bm{G}^\prime})-H(\bm{G}^\prime). 
    \end{aligned}
\end{equation}

By introducing a threshold $\tau_c$, non-uniform weights can have larger values of $P_c$, indicating the usage of VQ as shown in Figure~\ref{fig_proxy_methods}(a).  
Since IE is a measure of the entire system, a small amount of local outliers does not significantly effect $P_c$. 
However, in case of SQ, the accuracy is highly dependent to the data scale. Such outliers can cause more minimal values to be mapped to the same integer, thus increasing the rounding error. 
% For instance, the weight in Figure~\ref{fig_proxy_methods}(b) has close $P_c$ to that of Figure~\ref{fig_proxy_methods}(c), but its VQ result is better than SQ result. 
For instance, Figure~\ref{fig_proxy_methods}(b) and Figure~\ref{fig_proxy_methods}(c) have close $P_c$ values, while the former contains obvious outliers and is more accurate under VQ. 
\paragraph{Fine-grained Proxy}
% 细粒度指标
To mitigate the issue that $P_c$ is not sensitive enough to local outliers of a relatively uniform data, we further introduce a fine-grained proxy. 
Specifically, we perform the Taylor expansion~\cite{taylor1717methodus} to Equation~\ref{eq_coarse_proxy} to evaluate the minor disturbances $\bm{\delta}$ around $\hat{\bm{G}^\prime}$. 

\underline{\textit{Step 1}}~~The gap $\bm{\delta}$ between $\bm{G}^\prime$ and $\hat{\bm{G}^\prime}$ can be written as:
\begin{equation}
    \bm{\delta}=\bm{G}^\prime-\hat{\bm{G}^\prime}=\{\bm{\delta}_i=\bm{G}_i^\prime-\frac{1}{n}\thinspace|\thinspace i=1,2,...,n\}.
    \label{eq_delta}
\end{equation}
According to Equation~\ref{eq_interval_norm}, it should be satisfied that: 
\begin{equation}
    \sum_{i=1}^{n}\bm{\delta}_i=\sum_{i=1}^{n}\bm{G}_i^\prime-\sum_{i=1}^{n}\frac{1}{n}=0.
    \label{eq_sum_delta_is_zero}
\end{equation}
\underline{\textit{Step 2}}~~The Taylor expansion can be formulated as:
\begin{equation}
\begin{aligned}
    P_c(\bm{G}^\prime)&=P_c(\hat{\bm{G}^\prime})+\sum_{k=1}^{K}(k!)^{-1}P_c^k(\hat{\bm{G}^\prime})\bm{\delta}^k+o(\bm{\delta}^K)\\
    &=\sum_{k=1}^{K}(k!)^{-1}\sum_{i=1}^{n}\left.\frac{\partial^k P_c}{\partial \bm{G}_i^{\prime k}}\right|_{\bm{G}_i^\prime=\frac{1}{n}}\bm{\delta}_i^k+o(\bm{\delta}^K).
    \label{eq_taylor}
\end{aligned}
\end{equation}
Taking the Euler's number $e$ as the base of the $\log$ function in Equation~\ref{eq_coarse_proxy}, the $k$-th order partial derivative of $P_c$ with respect to $\bm{G}_i^\prime$ can be expressed as:
\begin{equation}
    \frac{\partial^k P_c}{\partial \bm{G}_i^{\prime k}}=\begin{cases}
  \ln \bm{G}_i^\prime+1 & k=1\\
  (-1)^k(k-2)!\bm{G}_i^{\prime (1-k)} & k\geq 2\end{cases}.
  \label{eq_partial}
\end{equation}
\underline{\textit{Step 3}}~~Taking Equation~\ref{eq_sum_delta_is_zero} and \ref{eq_partial} into consideration, Equation~\ref{eq_taylor} can be transformed into:
\begin{equation}
    P_c(\bm{G}^\prime)=\sum_{k=2}^{K}\frac{{(-1)}^kn^{k-1}}{k(k-1)}\sum_{i=1}^{n}\bm{\delta}_i^k+o(\bm{\delta}^K).
    \label{eq_taylor_final}
\end{equation}
\underline{\textit{Step 4}}~~Omitting the term $o(\bm{\delta}^K)$, Equation~\ref{eq_taylor_final} can be reformulated as: 
\begin{equation}
\begin{aligned}
    &P_c(\bm{G}^\prime)\approx [s_2, ..., s_K]\odot[v_2, ..., v_K]\odot[M_2, ..., M_K],\\
    &\mathrm{where}\quad s_k={(-1)}^k,\thinspace v_k=\frac{n^{k}}{k(k-1)}, \thinspace M_k=\frac{\sum_{i=1}^{n}\bm{\delta}_i^k}{n}.
    \label{eq_taylor_refomulation}
\end{aligned}
\end{equation}
Here, \textquoteleft$\odot$\textquoteright represents element-wise multiplication.  
$M_k$ is the $k$-th order central moment of $\bm{G}^\prime$, which is defined as:
\begin{equation}
    \begin{aligned}
        M_k(\bm{G}^\prime)&=\mathbb{E}[\thinspace {(\bm{G}^\prime-\mathbb{E}[\thinspace \bm{G}^\prime \thinspace])}^k \thinspace].
    \label{eq_central_moment_definition}
    \end{aligned}
\end{equation}
The central moment can serve as a metric for assessing the local features of data. 
This is because the difference between outliers and other data points is magnified by the $k$-th power.
For instance, when $k=2$, Equation~\ref{eq_central_moment_definition} yields the variance, indicating the spread of the data.
For $k=3$, the central moment corresponds to skewness, assessing the symmetry of the data.
For $k=4$, it represents kurtosis, revealing the data's long-tail characteristic. 

\underline{\textit{Step 5}}~~
Considering that $P_c(\bm{G}^\prime)$ signifies the overall uniformity, $s_k$ and $v_k$ can be regarded as the direction and the significance of the local feature $M_k$. Since only the magnitudes of features are considered when it comes to outliers, the fine-grained proxy can be defined as:
\begin{equation}
    P_f(\bm{G}^\prime)=\sum_{k=2}^{K}v_k|M_k|, 
    \label{eq_fine_proxy}
\end{equation}
% Since $0<G_i^\prime<1$, it should be satisfied that $-\frac{1}{n}<\delta_i<1-\frac{1}{n}(n>1)$. Thereby the absolute value of $\delta_i^k$ is expected to decrease as $k$ increases. In consideration of the computational cost, we opt to take the first four terms of Equation~\ref{eq_fine_proxy} in practice.
where $K$ is a hyper-parameter. 
By introducing a threshold $\tau_f$, outliers can be detected by larger values of $P_f$, indicating the usage of VQ as shown in Figure~\ref{fig_proxy_methods}(b). 

Finally, our proxy solution of Equation~\ref{eq_optimization_goal} can be obtained by the combination of $P_c$ and $P_f$:
\begin{equation}
\begin{aligned}
    \hat{\bm{\phi}}=\{&\phi_m=
    \begin{cases}
    1 &  P_c(\bm{G}_m^\prime)<\tau_c \thinspace \mathrm{and} \thinspace P_f(\bm{G}_m^\prime)<\tau_f\\
    0 & P_c(\bm{G}_m^\prime)<\tau_c \thinspace \mathrm{and} \thinspace P_f(\bm{G}_m^\prime)\geq \tau_f\\
     & \mathrm{or} \thinspace P_c(\bm{G}_m^\prime)\geq \tau_c \end{cases}
    \\
    &|\thinspace m=1,2,...,M\},
    \label{eq_multi_scale_proxy}
\end{aligned}
\end{equation}
where $\bm{G}_m^\prime$ denotes the $m$-th weight after the transformation. Only if both the coarse-grained proxy and the fine-grained proxy are lower than their corresponding threshold will SQ be applied, as shown in Figure~\ref{fig_proxy_methods}(c). Otherwise, the weight distribution is supposed to be generally uneven, or relatively uniform but with local outliers, which indicates the application of VQ. Notably, the fine-grained proxy is only utilized in condition that $P_c(\bm{G}_m^\prime) < \tau_c$. 

\input{tabs/overall_result}

\subsection{Codebook Optimization for Element-wise Multiplication}\label{subsec_coem}
Different from Transformer-based LLMs, the element-wise multiplication \textquoteleft$\odot$\textquoteright between the input $\bm{x}$ and the weight $\bm{\mu}$ is applied in all projection layers of the RWKV structure, as shown in Figure~\ref{fig_rwkv_struture}. 
In accordance with the proxy introduced in Section~\ref{subsec_hybrid}, VQ is expected to be applied to most of them. 
However, existing VQ methods are primarily tailored for matrix multiplication modules. 
We thereby propose to optimize the VQ codebook specifically for element-wise multiplication modules.

Given a weight $\bm{\mu}\in\mathbb{R}^{m\times n}$, it is first transformed to $\bm{\mu}^\prime\in\mathbb{R}^{(m\cdot n//d)\times d}$, where $d$ is the hidden dimension. 
Following the VQ process stated in Equation~\ref{eq_vq}, it can be quantized into $Q(\bm{\mu}^\prime)$. 
Typically, the quantization loss $\mathcal{L}$ can be written as:
% 调整公式上下间距
\begin{equation}
    \begin{aligned}
        \mathcal{L}&={||\bm{X}\odot\bm{\mu}^\prime-\bm{X}\odot \mathrm{Deq}(Q(\bm{\mu}^\prime))||}_F^2\\
        &=\sum_{i=1}^{m\cdot n//d}\sum_{j=1}^{d}\bm{X}_{ij}^2{(\Delta\bm{\mu}^\prime_{ij})}^2,
        \label{eq_quantization_loss_of_mul}
    \end{aligned}
\end{equation}
where $\bm{X}$ is a representative of the calibration activations, $\mathrm{Deq}$ is the de-quantization process, and $\Delta\bm{\mu}^\prime$ denotes the quantization error of the weight. 
To minimize Equation~\ref{eq_quantization_loss_of_mul}, a larger activation value should correspond to a smaller $\Delta\bm{\mu}^\prime$, indicating the significance of this position. 
Thus, we employ the term $\bm{X}^2$ to direct the weighted KMeans algorithm in the generalization of codebooks. 

Due to the nature of element-wise multiplication, $\bm{X}$ must have the same shape as $\bm{\mu}^\prime$, which further introduces an issue of integrating batches of data. 
The most straightforward approach is to simply average all samples. 
However, this method is not sufficiently effective because it is highly susceptible to the influence of a small number of outliers. 
Given that the activations of RWKV models typically follows an approximately normal distribution, we introduce a percentile-based clipping operation to limit the range of samples prior to averaging, thereby alleviating this issue. 

%% file: tabs/overall_result.tex
\begin{table*}[!t]
\renewcommand\arraystretch{1.1}
  \centering
  % \vspace{3mm}
  % 0-shot$^9$ includes ARC-easy, ARC-challenge, PIQA, and WinoGrande tasks, while 0-shot8adds BoolQ, SIQA, HellaSwag, LAMBADA(OpenAI) and OBQA tasks.
  \caption{Comparison of perplexity on LAMBADA and averaged accuracy on nine Zero-Shot tasks. For all methods except ours and floating-point, we report metrics under both bpw settings of 3.5 and 3.25.
  }
  %\vspace{-5mm}
  \label{overall_result}
  \setlength{\tabcolsep}{0.8mm}
  {
  \resizebox{\textwidth}{!}{
    \begin{tabular}{c|c|cc:cc:cc|cc:cc:cc:cc}
    % \toprule
    & & & & & & & & & & & & & \\
\noalign{\vspace{0.1em}}\hline\noalign{\vspace{0.1em}}
\hline\noalign{\vspace{0.1em}}
& & \multicolumn{2}{c:}{\textbf{RWKV7-0.1B}} & \multicolumn{2}{c|}{\textbf{RWKV7-0.5B}} & \multicolumn{2}{c:}{\textbf{RWKV7-1.47B}} & \multicolumn{2}{c:}{\textbf{RWKV6-1B}} &
\multicolumn{2}{c|}{\textbf{RWKV6-3B}} &
\multicolumn{2}{c:}{\textbf{RWKV6-7B}} &
\multicolumn{2}{c}{\textbf{RWKV6-14B}}
\\
\noalign{\vspace{0.1em}}\cdashline{3-16}\noalign{\vspace{0.1em}}
    % \midrule
    \textbf{Bpw.} & \textbf{Method} & 0-shot$^9$ & LambA.  & 0-shot$^9$ & LambA.  & 0-shot$^9$ & LambA.  & 0-shot$^9$ & LambA.  & 0-shot$^9$ & LambA.  & 0-shot$^9$ & LambA. & 0-shot$^9$ & LambA.  \\
     &       & Avg.($\uparrow$) & ($\downarrow$)   & Avg.($\uparrow$) & ($\downarrow$)   & Avg.($\uparrow$) & ($\downarrow$)   & Avg.($\uparrow$) & ($\downarrow$)   & Avg.($\uparrow$) & ($\downarrow$)   & Avg.($\uparrow$) & ($\downarrow$)   & Avg.($\uparrow$) & ($\downarrow$)   \\
    % \midrule
    \noalign{\vspace{0.1em}}\hdashline\noalign{\vspace{0.1em}} 
    16 & FloatingPoint & 43.02  & 14.21 & 48.67  & 7.21  & 55.08  & 4.80  & 54.39  & 4.60  & 58.32  & 3.83 & 61.69  & 3.21  & 63.65  & 3.02
    \\
    % \midrule
\noalign{\vspace{0.1em}}\hdashline\noalign{\vspace{0.1em}} 
\multirow{7}[2]{*}{3.25} 
          & RTN    & 36.22  & 152.82  & 39.99  & 57.11 & 45.46  & 11.43  & 49.84  & 6.39  & 54.17  & 4.71  & 58.34  & 3.87  & 61.16  & 3.34   \\
          & GPTQ   & 37.92  & 63.54   & 41.16  & 23.29 & 51.15  & 7.93   & 50.55  & 6.43  & 53.94  & 4.88  & 59.28  & 3.72  & 60.18  & 3.43   \\
          & AWQ    & 36.20  & 132.06  & 68.92  & 5.92  & 43.62  & 15.27  & 44.41  & 17.97 & 47.24  & 11.97 & 49.48  & 8.33  & 49.35  & 8.18   \\          
          & QuaRot & 34.53  & 243.99  & 40.17  & 76.89 & 50.81  & 9.39   & 45.02  & 29.38 & 48.24  & 22.67 & 54.29  & 8.81  & 54.76  & 14.05  \\
          & kMeans & 38.21  & 87.06   & 44.59  & 20.19 & 52.77  & 6.57   & 47.02  & 15.93 & 53.06  & 8.27  & 55.72  & 4.69  & 61.40  & 4.61   \\
          & GPTVQ  & 40.25  & 23.75   & 43.64  & 14.15 & 52.06  & 5.54   & 49.86  & 6.11  & 54.23  & 4.31  & 58.57  & 3.49  & 59.63  & 3.15   \\
          & VPTQ   & 35.78  & 128.59  & 40.14  & 30.63 & 45.83  & 11.13  & 43.89  & 14.67 & 48.22  & 7.77  & 53.06  & 4.75  & 57.87  & 3.62   \\
\noalign{\vspace{0.1em}}\hline\noalign{\vspace{0.1em}}
\multirow{7}[2]{*}{3.5} 
          & RTN    & 38.08  & 81.14  & 43.02  & 25.09  & 51.74  & 7.89  & \textbf{51.76}  & 5.83  & 54.42  & 4.50   & 59.20  & 3.59  & 60.84  & 3.31    \\
          & GPTQ   & 39.33  & 40.16  & 45.73  & 13.07  & 52.28  & 6.55  & 51.23  & 5.86  & 54.89  & 4.54  & 60.07  & 3.68  & 61.14  & 3.29    \\
          & AWQ    & 38.31  & 55.72  & 42.40  & 16.98  & 44.61  & 10.71 & 51.20  & 10.07 & 48.40  & 8.77  & 50.50  & 7.07  & 50.86  & 6.06    \\
          & QuaRot & 37.26  & 126.19 & 40.84  & 40.38  & 51.98  & 7.94  & 47.00  & 16.29 & 51.01  & 16.99 & 56.95  & 6.44  & 56.33  & 10.63   \\
          & kMeans & 39.55  & 36.26  & 43.07  & 17.05  & \textbf{52.77}  & 6.57  & 50.03  & 8.24  & 54.99  & 6.28  & 59.51  & 3.96  & 61.15  & 3.84    \\
          & GPTVQ  & 40.10  & 25.82  & 44.13  & 10.88  & 52.13  & 5.51  & 50.29  & 5.74  & 55.12  & 4.12  & 59.70  & 3.30  & 59.76  & 3.34    \\
          & VPTQ   & 37.07  & 74.70  & 41.06  & 25.03  & 47.38  & 9.52  & 43.82  & 14.74 & 48.86  & 8.62  & 52.95  & 4.47  & 57.93  & 3.75    \\
          
          %& \cellcolor[rgb]{ .906,  .902,  .902}\textbf{OSTQuant} & \cellcolor[rgb]{ .906,  .902,  .902}\textbf{67.80} & \cellcolor[rgb]{ .906,  .902,  .902}6.53  & \cellcolor[rgb]{ .906,  .902,  .902}\textbf{73.69} & \cellcolor[rgb]{ .906,  .902,  .902}\textbf{3.19} & \cellcolor[rgb]{ .906,  .902,  .902}\textbf{64.37} & \cellcolor[rgb]{ .906,  .902,  .902}5.64  & \cellcolor[rgb]{ .906,  .902,  .902}\textbf{67.31} & \cellcolor[rgb]{ .906,  .902,  .902}\textbf{4.94} & \cellcolor[rgb]{ .906,  .902,  .902}\textbf{71.48} & \cellcolor[rgb]{ .906,  .902,  .902}\textbf{3.41}  & \cellcolor[rgb]{ .906,  .902,  .902}\textbf{64.13} & \cellcolor[rgb]{ .906,  .902,  .902}5.81  & \cellcolor[rgb]{ .906,  .902,  .902}\textbf{66.62} & \cellcolor[rgb]{ .906,  .902,  .902}5.21    \\
          
\noalign{\vspace{0.1em}}\hline\noalign{\vspace{0.1em}}
\multirow{1}[1]{*}{3.275}   &\cellcolor[rgb]{ .906,  .902,  .902}\textbf{Ours} & \cellcolor[rgb]{ .906,  .902,  .902}\textbf{41.10} & \cellcolor[rgb]{ .906,  .902,  .902}\textbf{18.41}  & \cellcolor[rgb]{ .906,  .902,  .902}\textbf{46.01} & \cellcolor[rgb]{ .906,  .902,  .902}\textbf{9.39} & \cellcolor[rgb]{ .906,  .902,  .902}52.40 & \cellcolor[rgb]{ .906,  .902,  .902}\textbf{5.24}  & \cellcolor[rgb]{ .906,  .902,  .902}51.69 & \cellcolor[rgb]{ .906,  .902,  .902}\textbf{5.29} & \cellcolor[rgb]{ .906,  .902,  .902}\textbf{55.79} & \cellcolor[rgb]{ .906,  .902,  .902}\textbf{3.88}  & \cellcolor[rgb]{ .906,  .902,  .902}\textbf{60.19} & \cellcolor[rgb]{ .906,  .902,  .902}\textbf{3.23  } & \cellcolor[rgb]{ .906,  .902,  .902}\textbf{62.69} & \cellcolor[rgb]{ .906,  .902,  .902}\textbf{2.89}    \\

    % \bottomrule
    \noalign{\vspace{0.1em}}\hline\noalign{\vspace{0.1em}}
\hline\noalign{\vspace{0.1em}}
    \end{tabular}%
    }
    }
  %\vspace{-5mm}
\end{table*}%

%% file: texs/5_experiments.tex
\section{Experiments}\label{sec_experiments}
\subsection{Experimental Settings}\label{subsec_experimental_settings}
\textbf{Models and Datasets.} We evaluate the RWKVQuant framework on RWKV6~\cite{peng2024eagle}, RWKV7, and VRWKV models~\cite{duan2024vision}. 
% We evaluate the performance of the quantized RWKV models across vision and language tasks. 
For vision tasks, we utilize ImageNet~\cite{deng2009imagenet} for image classification, Coco~\cite{lin2014microsoft} for object detection, and ADE20K~\cite{zhou2019semantic} for segmentation. 
Aligned with the accuracy evaluation methods used in the VRWKV experiments, we report Top-1 Accuracy for classification tasks, Box Average Precision (AP) for detection tasks, and Mean Intersection over Union (MIoU) for segmentation tasks. 
For language tasks, consistent with the RWKV6 paper, we report the perplexity (PPL) on the Lambada dataset. We also evaluate the models on up to nine zero-shot tasks using the LM-evaluation-harness (version 0.4.4), including LAMBADA(OpenAI) ~\cite{radford2019language}, HEADQA (EN) ~\cite{rogers2023qa}, HellaSwag ~\cite{zellers2019hellaswag}, OpenBookQA (OBQA) ~\cite{mihaylov2018can}, PIQA ~\cite{bisk2020piqa}, SCIQ ~\cite{pedersen2020sciq}, Winogrande~\cite{sakaguchi2021winogrande}, ARC-Challenge and ARC-Easy ~\cite{boratko2018systematic}.

\textbf{Baselines and Implementation Details.} In addition to comparing with SQ methods such as RTN, GPTQ~\cite{frantar2022gptq}, AWQ~\cite{lin2023awq}, and Qurot~\cite{ashkboos2024quarot}, we also benchmark our approach against VQ methods like K-Means, GPTVQ~\cite{van2024gptvq} and VPTQ~\cite{liu2024vptq} for weight-only quantization. To ensure fairness, we report the performance of each method under two configurations, where the average number of bits per weight (bpw) is set to 3.25 and 3.5. For SQ methods, we take the scale size into account when calculating the bpw. To achieve 3.25 and 3.5 bits per weight, we set the group size for quantization to 32 and 64 respectively. For VQ methods, we consider not only the bit size occupied by the quantized weights but also the bit size required for storing the codebook to achieve the corresponding bpw. In our method, we dynamically set $\tau_c$ and $\tau_f$ according to different models, ensuring that SQ with a bpw of 3.25 is used in nine-tenths of the layers, while VQ with a bpw of 3.5 is used in one-tenth. For example, in RWKV7, $\tau_c$ is set to 1.54, while $\tau_f$ is set to 30. For both vision and language tasks, we select 128 samples from the corresponding test datasets for calibration.

\subsection{Overall Results}\label{subsec_overall_res}
\textbf{Performance Comparison on Language Tasks.} As shown in Table~\ref{overall_result}, on language tasks our method consistently outperforms other approaches across nearly all models. Compared to methods with a bpw of 3.25, regardless of whether they based on SQ or VQ, our method demonstrates significant improvements in both PPL and accuracy on zero-shot tasks. Compared to methods with a bpw of 3.5, our method consistently achieves lower PPL. Except for slightly lower accuracy on RWKV7-0.5B and RWKV6-1B with certain methods, it achieves the highest accuracy across all other models. It can be observed that on the smallest 0.1B model, the PPL of other methods increases by at least 10 points, whereas our method results in an increase of only 4.2 points. On larger models such as RWKV6-7B and RWKV7-14B, our method results in almost no increase in PPL, while the accuracy decreases by less than 1 point. 

\input{tabs/overall_visual_res}
\textbf{Performance Comparison on Vision Tasks.} Table~\ref{overall_visual_res} presents the results of the quantized RWKV models applied to various vision tasks, including classification, detection and segmentation. 
% The table compares the performance of several quantization methods, including GPTQ, AWQ, GPTVQ, VPTQ and our proposed method, across different RWKV model variants. Other methods use configurations with a bpw of 3.5, while our method maintains a bpw of 3.275. 
Our method achieves the highest scores in both segmentation and classification tasks. For detection tasks, although the precision of RWKV-S is not the highest, it is very close to the best-performing method. 
% All results for vision and language models under our method are in the Appendix.

% \input{tabs/overall_visual_res}
\input{tabs/speed}

\textbf{Memory Occupancy and Computational Cost.} Our method incurs only negligible loss in 3.275-bpw quantization, making 3.275-bpw inference feasible. As described in the section \ref{sec_intro}, models based on the RWKV architecture differ from those built on GPT or LLaMA architectures. Whether in the pre-fill or decoder stage, RWKV models exhibit a lower compute-to-memory-access ratio. Consequently, quantizing the weights to lower bit-widths can significantly reduce memory access time, thereby accelerating the model's inference speed, as shown in Table~\ref{Tab speedup}.

\input{tabs/abliation_mix_res}

\subsection{Ablation Study}
\textbf{Hybrid Quantization.} We conduct a series of ablation studies on the hybrid quantization method proposed in Section \ref{subsec_hybrid}, comparing its performance on the RWKV model with that of employing single quantization methods.
For fairness, the weights of all multiplication operations are quantized using the RTN method. Our method leverages the proposed coarse-grained and fine-grained proxy to hybridize GPTQ and GPTVQ. While GPTQ and GPTVQ use a bpw of 3.5, our method achieves a bpw of 3.275 by applying GPTVQ (bpw 3.5) to one-tenth of the layers and GPTQ (bpw 3.25) to the remaining nine-tenths. The ablation study results in Table~\ref{abliation_mix_res} highlight the effectiveness of the hybrid quantization method. In nearly all RWKV models, the hybrid method achieves better metrics compared to both GPTQ and GPTVQ.

\input{tabs/abliation_proxy}

\textbf{Proxy Strategy.} Table~\ref{abliation_mix_res}  shows that our hybrid quantization improves accuracy but still lags behind floating-point precision. 
We then apply the proxy described in Section~\ref{subsec_hybrid}, combining coarse-grained and fine-grained proxies to determine the quantization method for each layer. The ablation results, comparing the use of different proxies such as Variance, Coefficient of Variation (CV) ~\cite{abdi2010coefficient}, Range~\cite{10.1214/aoms/1177730387}, Mean Absolute Deviation (MAD)~\cite{konno2005mean}, Mean Squared Error(MSE), and IE, are presented in Table~\ref{abliation_proxy}. 
% The use of MSE involves calculating the MSE loss of the quantized weights for each layer after applying SQ and VQ quantization to the weights, and then selecting the method with the lower loss. 
Notably, MSE denotes making selections between SQ and VQ by directly comparing their MSE of each weight. 
The other metrics are used in the same manner as described in our method, focusing on the transformed weights $G^\prime$. 
Intuitively, the MSE method is the local optimum for each weight. 
However, our coarse-to-fine proxy attains the best results across all three models from the global perspective. 
\input{tabs/abliation_mul_res}
\begin{figure}[!b]
    \centering
    \includegraphics[width=0.48\textwidth]{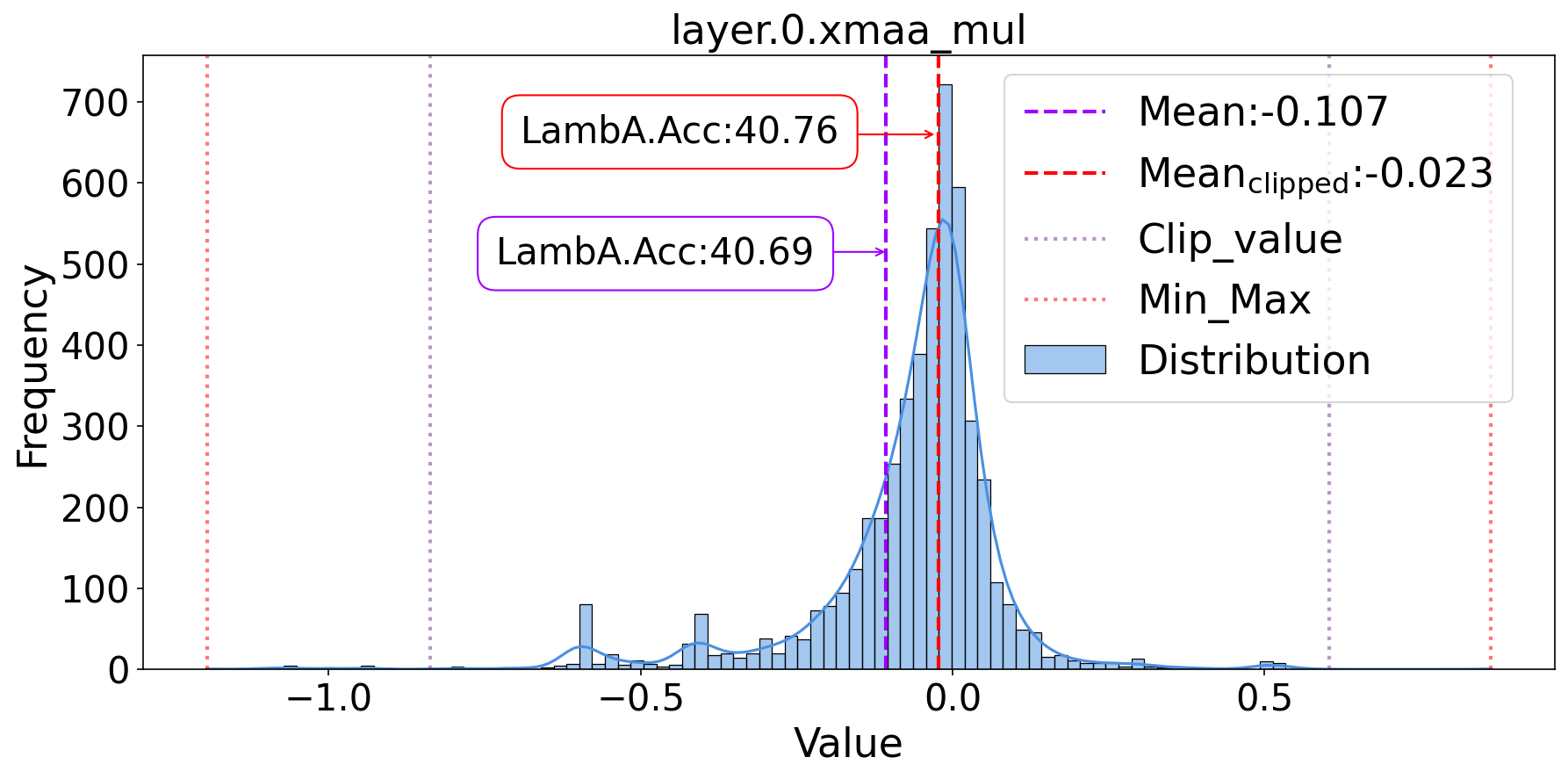}
    \caption{Effectiveness of clipping for batch integration.}
    \label{fig_mul_input}
\end{figure}

\textbf{Codebook Optimization.} We conduct ablation experiments on the codebook optimization for the element-wise
multiplication proposed in Section~\ref{subsec_coem}, across all RWKV models. The results are presented in Table~\ref{abliation_mul_res}. 
It can be observed that using the codebook optimization for the element-wise multiplication operator generates better accuracy across all models compared to not applying the optimization. 
Specifically, we also visualize the effectiveness of the clipping-based percentile technique within this codebook optimization. 
From Figure~\ref{fig_mul_input}, it can be clearly observed that the input activation approximately follows the normal distribution. 
However, the outliers make the representative feature to leave far from the center point, thereby decreasing the overall performance. 
By clipping these outliers, a more close-to-center feature can be obtained, thus enhancing the calibration process.

\subsection{More Uniform Weights in RWKV}\label{subsec_analysis_rwkv_uniformity}
\begin{figure}[!t]
    \centering
    \includegraphics[width=0.4\textwidth]{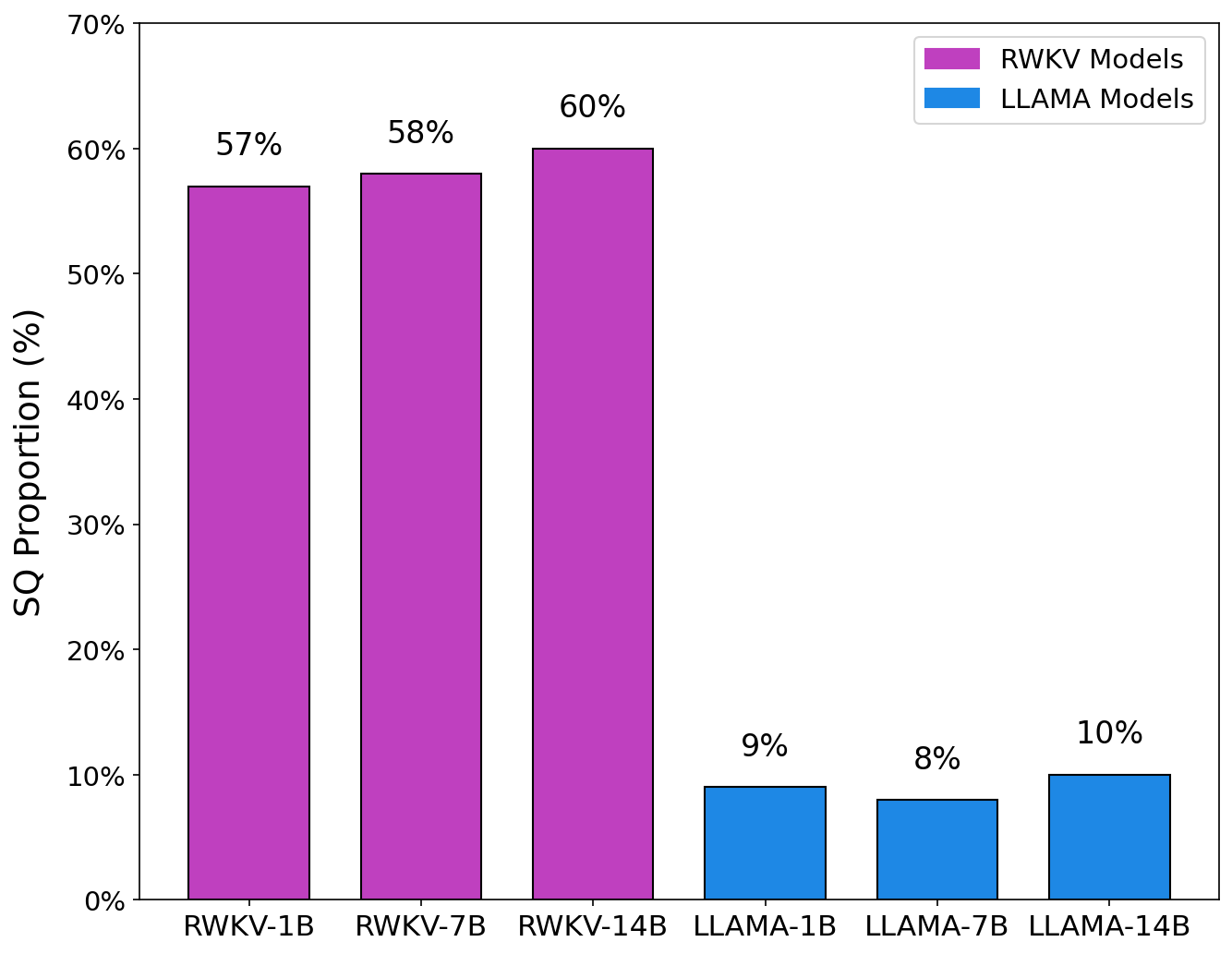}
    \caption{Comparison of SQ proportion between RWKV and LLaMA Models.}
    \label{SQ_percent}
    \vspace{-3 mm}
\end{figure}
Table~\ref{tab_kmeans_loss} in Section~\ref{sec_intro} presents the average relative clustering loss of weights using K-Means methods~\cite{lloyd1982least} for the RWKV family and the LLaMA family respectively. 
In depth, we conduct experiments leveraging the proposed coarse-to-fine proxy in Section~\ref{subsec_hybrid} to investigate the usage proportions of SQ and VQ.
Under the settings of $\tau_c=1.5$ and $\tau_f=50$, Figure~\ref{SQ_percent} shows that approximately 60\% of the layers in the RWKV family are categorized as suitable for scaler quantization, whereas the proportion is only about 10\% for the LLaMA family. This further demonstrates that the RWKV models have a significantly higher number of uniform weights.

%% file: tabs/overall_visual_res.tex
\begin{table}[!t]
\renewcommand\arraystretch{1.2} % 调整行高
\centering
% \vspace{-2mm}
\caption{Comparative results under different quantization settings for Vision RWKV models.}
% \vspace{-2mm}
\label{overall_visual_res}
\setlength{\tabcolsep}{0.5mm} % 调整列间距
\resizebox{\columnwidth}{!}{
\begin{tabular}{c|c|ccc|ccc}
\hline
\multirow{2}{*}{\textbf{Bpw.}} & \multirow{2}{*}{\textbf{Method}} & \multicolumn{3}{c|}{\textbf{RWKV-T}} & \multicolumn{3}{c}{\textbf{RWKV-S}} \\
\noalign{\vspace{0.1em}}\cdashline{3-8}\noalign{\vspace{0.1em}}
             &                 & \textbf{Cls.} & \textbf{Det.} & \textbf{Seg.} & \textbf{Cls.} & \textbf{Det.} & \textbf{Seg.} \\
\hline
16       & FloatingPoint      & 75.10 & 41.70 & 43.3 & 80.10 & 44.8 & 47.2 \\
\hline
\multirow{4}{*}{3.5} 
             & GPTQ           & 69.74  & 39.85 & 41.20 & 78.30 & 43.37 & 45.50 \\
             & AWQ            & 68.50  & 39.03 & 38.88 & 78.00 & 42.90 & 42.88 \\          
             & GPTVQ          & 70.31  & 40.14 & 41.65 & 78.65 & \textbf{44.03} & 45.00 \\
             & VPTQ           & 67.21  & 39.02 & 40.14 & 76.40 & 42.01 & 43.54 \\
\hline
3.275        & \cellcolor[rgb]{ .906,  .902,  .902}\textbf{Ours} & \cellcolor[rgb]{ .906,  .902,  .902}\textbf{70.41}  & \cellcolor[rgb]{ .906,  .902,  .902}\textbf{40.22} & \cellcolor[rgb]{ .906,  .902,  .902}\textbf{41.70} & \cellcolor[rgb]{ .906,  .902,  .902}\textbf{78.74} & \cellcolor[rgb]{ .906,  .902,  .902}43.95 & \cellcolor[rgb]{ .906,  .902,  .902}\textbf{46.09} \\
\hline
\end{tabular}%
 }
% \vspace{-3mm}
\end{table}

%% file: tabs/speed.tex
% \vspace{-20mm}
\begin{table}[!b]
%\vspace{-6mm}
\renewcommand\arraystretch{1.3} % 调整行高
\setlength{\tabcolsep}{0.5mm} % 调整列间距
\centering
\caption{Comparison of generation speed and memory usage before and after 3.275-bpw quantization on RWKV6 models. All tests were conducted on an NVIDIA A6000 GPU.}
%\vspace{-6mm}
\label{Tab speedup}
\resizebox{\columnwidth}{!}{
\begin{tabular}{cccc|ccc}
& & & & & \\
\toprule
\multirow{2}{*}{\textbf{Model Size}} & \multicolumn{3}{c|}{\textbf{ speed (tokens/sec)}} & \multicolumn{3}{c}{\textbf{Memory use (GB)	}} \\
\cmidrule{2-7} \
 & \textbf{FP} & \textbf{Quantized} & \textbf{Speed up}  & \textbf{FP} & \textbf{Quantized} & \textbf{Mem. saving}\\
\noalign{\vspace{1mm}}\hline\noalign{\vspace{1mm}}
    3B  & 32.95 & 51.29 & 1.55x & 5.88  & 1.65 & 3.56x \\
    7B  & 30.75 & 62.42 & 2.03x & 13.91 & 4.25 & 3.27x \\
    14B & 16.02 & 34.32 & 2.14x & 26.07 & 9.21 & 2.83x \\
 \bottomrule
\end{tabular}}
%\vspace{-3mm}
\end{table}

%% file: tabs/abliation_mix_res.tex
\begin{table}[!t]
\renewcommand\arraystretch{1.2} % 调整行高
\centering
% \vspace{-1mm}
\caption{Ablation study on the impact of hybrid quantization on LAMBADA PPL and zero-shot$^9$ score for language RWKV models.}
% \vspace{-2mm}
\label{abliation_mix_res}
\setlength{\tabcolsep}{0.1mm} % 调整列间距
\resizebox{\columnwidth}{!}{%
\begin{tabular}{c|cc|cc|cc}
\hline
  \multirow{3}{*}{\textbf{Model}} & \multicolumn{2}{c|}{\textbf{GPTQ}} & \multicolumn{2}{c|}{\textbf{GPTVQ}} & \multicolumn{2}{c}{\textbf{Ours}}\\
\noalign{\vspace{0.1em}}\cdashline{2-7}\noalign{\vspace{0.1em}}
             & 0-shot$^9$ & LambA.  & 0-shot$^9$ & LambA.  & 0-shot$^9$ & LambA. \\
             & Avg.($\uparrow$) & ($\downarrow$)   & Avg.($\uparrow$) & ($\downarrow$)   & Avg.($\uparrow$) & ($\downarrow$)\\
\hline
        RWKV7-0.1B      & 39.33  & 40.16  & 38.49  & 55.30   & 40.69  & 24.71 \\
        RWKV7-0.5B      & 45.36  & 13.07  & 43.85  & 20.16   & 45.03  & 13.49 \\
        RWKV7-1.47B     & 52.28  & 6.55   & 51.31  & 6.85    & 52.23  & 6.54 \\   
        RWKV6-1B        & 51.20  & 5.86   & 49.70  & 5.52    & 51.44  & 5.32 \\
        RWKV6-3B        & 55.24  & 4.54   & 54.86  & 4.41    & 55.40   & 3.97 \\
        RWKV6-7B        & 59.20  & 3.59   & 48.29  & 3.41    & 60.18  & 3.21 \\
        RWKV6-14B       & 61.14  & 3.29   & 59.86  & 3.31    & 62.03  & 2.89 \\

\hline
\end{tabular}%
}
% \vspace{-7.5mm}
\end{table}

%% file: tabs/abliation_proxy.tex
\begin{table}[!b]
\renewcommand\arraystretch{1} % 调整行高
\centering
%\vspace{-1mm}
\caption{Ablation study on the impact of different proxies for hybrid quantization in language RWKV models.}
%\vspace{-3mm}
\label{abliation_proxy}
\setlength{\tabcolsep}{0.1mm} % 调整列间距
\resizebox{\columnwidth}{!}{%
\begin{tabular}{c|cc|cc|cc}
\hline
  \multirow{3}{*}{\textbf{Method}} & \multicolumn{2}{c|}{\textbf{RWK7-0.1B}} & \multicolumn{2}{c|}{\textbf{RWK7-0.5B}} & \multicolumn{2}{c}{\textbf{RWK7-1.47B}}\\
\noalign{\vspace{0.1em}}\cdashline{2-7}\noalign{\vspace{0.1em}}
             & 0-shot$^9$ & LambA.  & 0-shot$^9$ & LambA.  & 0-shot$^9$ & LambA. \\
             & Avg.($\uparrow$) & ($\downarrow$)   & Avg.($\uparrow$) & ($\downarrow$)   & Avg.($\uparrow$) & ($\downarrow$)\\
\hline
Variance  & 40.67 & 20.80 & 42.51 & 9.74 & 51.29 & 5.90 \\
CV        & 39.09 & 22.36 & 40.92 & 10.10 & 51.58 & 5.79 \\
Range     & 39.92 & 23.78 & 40.24 & 10.41 & 51.37 & 5.82 \\
MAD       & 38.97 & 22.65 & 42.33 & 10.02 & 51.95 & 6.04 \\
MSE       & 37.99 & 28.56 & 42.60 & 10.22 & 51.05 & 6.87 \\
IE        & 41.01 & 20.03 & 45.12 & 9.67 & 52.12 & 5.31 \\
\textbf{Ours}      & \cellcolor[rgb]{ .906,  .902,  .902}\textbf{41.04} & \cellcolor[rgb]{ .906,  .902,  .902}\textbf{19.70} & \cellcolor[rgb]{ .906,  .902,  .902}\textbf{45.54} & \cellcolor[rgb]{ .906,  .902,  .902}\textbf{9.55} & \cellcolor[rgb]{ .906,  .902,  .902}\textbf{52.32} & \cellcolor[rgb]{ .906,  .902,  .902}\textbf{5.24} \\
\hline
\end{tabular}%
}
%\vspace{-4mm}
\end{table}

%% file: tabs/abliation_mul_res.tex
\begin{table}[!t]
\renewcommand\arraystretch{1.0} % 调整行高
\centering
\small
%\vspace{-1mm}
\caption{Ablation study on the impact of codebook optimization for element-wise
multiplication.}
%\vspace{-3mm}
\label{abliation_mul_res}
\setlength{\tabcolsep}{0.8mm} % 调整列间距
\scalebox{0.9}{
\resizebox{\columnwidth}{!}{
\begin{tabular}{c|cc|cc}
\hline
  \multirow{3}{*}{\textbf{Model}} & \multicolumn{2}{c|}{\textbf{wo.}} & \multicolumn{2}{c}{\textbf{w.}} \\
\noalign{\vspace{0.1em}}\cdashline{2-5}\noalign{\vspace{0.1em}}
             & 0-shot$^9$ & Lambda & 0-shot$^9$ & Lambda \\
             & Avg.($\uparrow$) & ($\downarrow$)   & Avg.($\uparrow$) & ($\downarrow$)   \\
\noalign{\vspace{0.1em}}\hline\noalign{\vspace{0.1em}}\

        RWKV7-0.1B      & 40.69  & 24.71  & 41.09  & 18.41   \\
        RWKV7-0.5B      & 45.03  & 13.49  & 46.01  & 9.39   \\
        RWKV7-1.47B     & 52.24  & 6.54   & 52.41  & 5.24    \\
        RWKV6-1B        & 51.45  & 5.32   & 51.69  & 5.29    \\
        RWKV6-3B        & 55.48  & 3.97   & 55.79  & 3.88    \\
        RWKV6-7B        & 60.18  & 3.21   & 60.19  & 3.23    \\
        RWKV6-14B       & 62.03  & 2.89   & 62.69  & 2.89    \\
\hline
\end{tabular}%
}
}
%\vspace{-5mm}
\end{table}

%% file: texs/6_conclusion.tex
\section{Conclusion}\label{sec_conclusion}
In this paper, we focus on introducing the quantization techniques into the realm of RWKV models. 
% Firstly, we find that parameters of recently advanced SQ approaches, i.e., smooth- and rotation-based methods, can not be fused into weights. 
% This limitation inevitably results in additional runtime computation.
% Secondly, we observe that VQ, which is based on clustering, suffers obvious decline of accuracy. 
% This is primarily owing to the larger amount of uniform weights in RWKV that posing challenges to the clustering process. 
Our investigation reveals that applying SQ or VQ individually may not be optimal for RWKV. We have subsequently identified that enhancing VQ with conventional compensation-based SQ holds great promise.
To this end, we propose RWKVQuant, a comprehensive post training quantization framework especially designed for RWKV models.  
The core idea is to design an optimal strategy that indicates the choice between SQ and VQ for each weight.
Specifically, we propose a guidance that employs a coarse-grained proxy to evaluate uniformity and a fine-grained proxy to identify outliers. 
We also optimize the codebook generation for element-wise multiplication modules, which are unique to the RWKV models. 
Our proposed RWKVQuant advances in accuracy for both RWKV-based vision and language tasks compared to existing methods, making RWKV models more practical for deployment in resource-constrained environments. 
As a pioneering study on quantization within the RWKV family,
we will publish the code in the hope of promoting further research and facilitating advancements in this field.

%% file: texs/7_appendix.tex
\clearpage
\onecolumn
\appendix
\section{Appendix}
\label{appendix}

\subsection{Structure of Time- and Channel-Mixing}\label{appendix_rwkv_structure}
The RWKV model, similar to Transformer networks, is composed of multiple identical blocks, each containing a Time Mixing component. 
% Within these components, the RWKV model incorporates numerous nonlinear operators and Element-wise Multiplication. These nonlinear operators are a key factor contributing to the additional computational overhead observed with methods like Smooth and Qurot. 
The Time Mixing process can be written as:
 \begin{equation} %这里看篇幅大小，适当压缩成两个公式。类似self-attention的方式。% OK，这里先暂时用的RWKV4原文的公式，后续看情况调整一下。
     \bm{r}_t = \bm{W}_r \cdot (\bm{\mu}_r \odot \bm{x}_t + (1 - \bm{\mu}_r) \odot \bm{x}_{t-1}),
     \label{eq_time_mixing_r}
 \end{equation}
 \begin{equation}
     \bm{k}_t = \bm{W}_k \cdot (\bm{\mu}_k \odot \bm{x}_t + (1 - \bm{\mu}_k) \odot \bm{x}_{t-1}),
     \label{eq_time_mixing_k}
 \end{equation}
 \begin{equation}
     \bm{v}_t = \bm{W}_v \cdot (\bm{\mu}_v \odot \bm{x}_t + (1 - \bm{\mu}_v) \odot \bm{x}_{t-1}),
     \label{eq_time_mixing_v}
 \end{equation}
 \begin{equation}
 \label{eq_time_mixing_wkv}
     \begin{aligned}
         \bm{wkv}_t=\frac{\sum_{i = 1}^{t - 1} f(i) \odot \bm{v}_i + \exp(\bm{u} + \bm{k}_t) \odot \bm{v}_t}{\sum_{i = 1}^{t - 1} f(i) +\exp(\bm{u} + \bm{k}_t)} \\
         where f(i) = \exp(- (t - 1 - i)\bm{w} + \bm{k}_i),
     \end{aligned}
 \end{equation}
 \begin{equation}
     \bm{o}_t = \bm{W}_o \cdot (\sigma(\bm{r}_t) \odot \bm{wkv}_t).
     \label{eq_time_mixing_o}
 \end{equation}

Here, the symbol \textquoteleft$\odot$\textquoteright represents element-wise multiplication, while the symbol \textquoteleft$\thinspace\cdot\thinspace$\textquoteright stands for matrix multiplication. Both $W$ and $\mu$ are parameters. In the context of RWKV, the terms $r_t$, $k_t$, and $v_t$ bear an analogy to the Q, K, and V components found in the attention mechanism of Transformers. Notably, the input x in RWKV is not simply the embedding of the current token. Rather, it signifies the weighted sum of the embedding of the current token and that of the previous token. 
Subsequently, the Channel Mixing module performs: 
\begin{equation}
    \bm{r}_t^\prime = \bm{W}_r^\prime \cdot (\bm{\mu}_r^\prime \odot \bm{x}_t + (1 - \bm{\mu}_r^\prime) \odot \bm{x}_{t-1}),
    \label{eq_channel_mixing_r}
\end{equation}
\begin{equation}
    \bm{k}_t^\prime = \bm{W}_k^\prime \cdot (\bm{\mu}_k^\prime \odot \bm{x}_t + (1 - \bm{\mu}_k^\prime) \odot \bm{x}_{t-1}),
    \label{eq_channel_mixing_k}
\end{equation}
\begin{equation}
    \bm{o}_t^\prime = \sigma(\bm{r}_t^\prime) \odot (\bm{W}_v^\prime \cdot \max (\bm{k}_t^\prime, 0)^2).
    \label{eq_channel_mixing_o}
\end{equation}

\subsection{RWKV Weight Distribution}\label{rwkv_w_distribute}
\begin{figure}[h]
    \centering
    \includegraphics[width=0.98\textwidth]{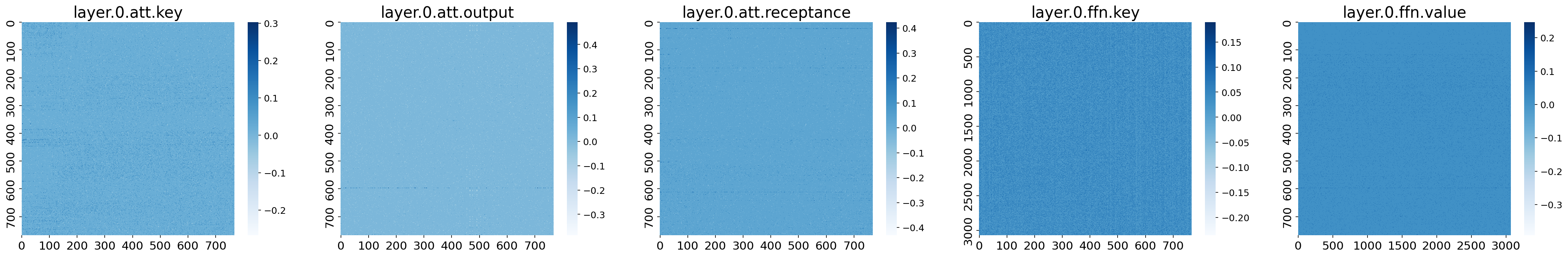}
    \caption{Unifrom weights without outliers in RWKV7-0.1B different layers.}
    \label{rwkv_w_unif}
\vspace{-5mm}
\end{figure}

\begin{figure}[h]
    \centering
    \includegraphics[width=0.98\textwidth]{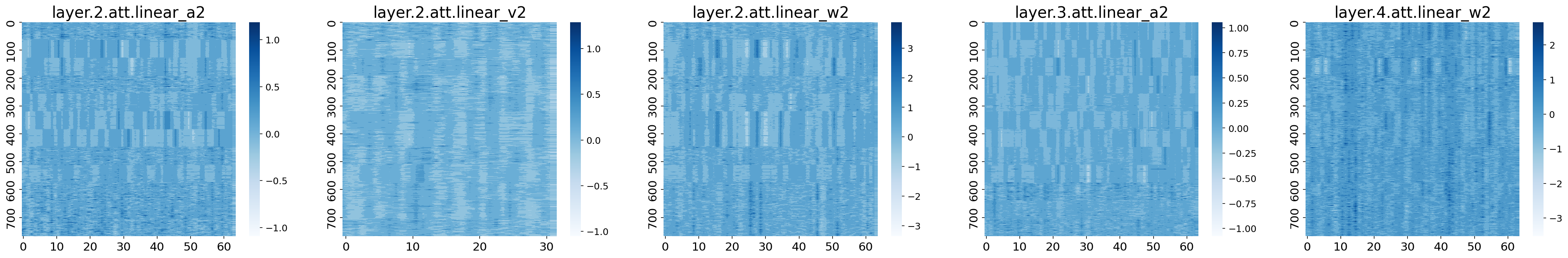}
    \caption{Non-uniform weights in RWKV7-0.1B different layers.}
    \label{rwkv_w_Non-unif}
\vspace{-5mm}
\end{figure}

\begin{figure}[t]
    \centering
    \includegraphics[width=0.98\textwidth]{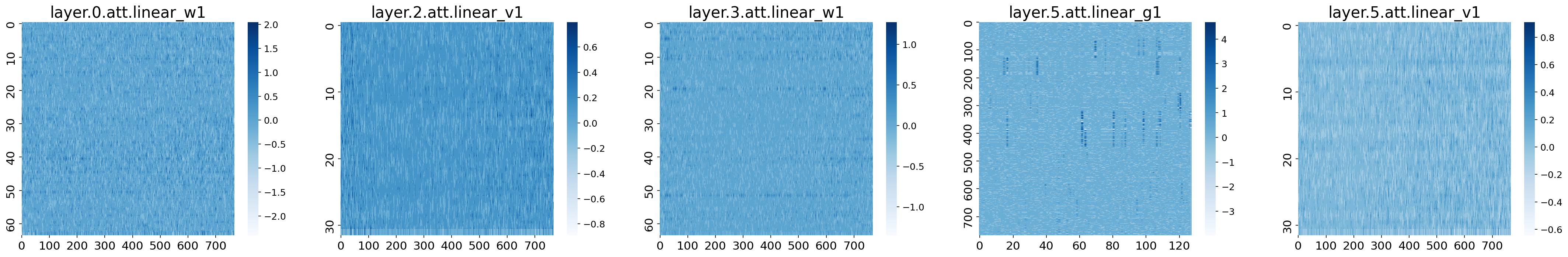}
    \caption{Unifrom weights with outliers in RWKV7-0.1B different layers.}
    \label{rwkv_w_local}
\vspace{-5mm}
\end{figure}

Figue~\ref{rwkv_w_unif} shows layers with relatively uniform weight distributions in the RWKV7-0.1B model, which are classified as layers that should use SQ based on our proposed coarse-to-fine proxy. In contrast, Figure~\ref{rwkv_w_Non-unif} illustrates layers with uneven weight distributions, which are typically classified as layers that should use VQ. Furthermore, although the weights in Figure~\ref{rwkv_w_local} appear generally uniform, the presence of local unevenness still leads to their classification as layers that require VQ.

\subsection{Compute-to-memory Ratio}\label{appendix a}
\begin{figure}[H]
    \centering
    \includegraphics[width=0.95\textwidth]{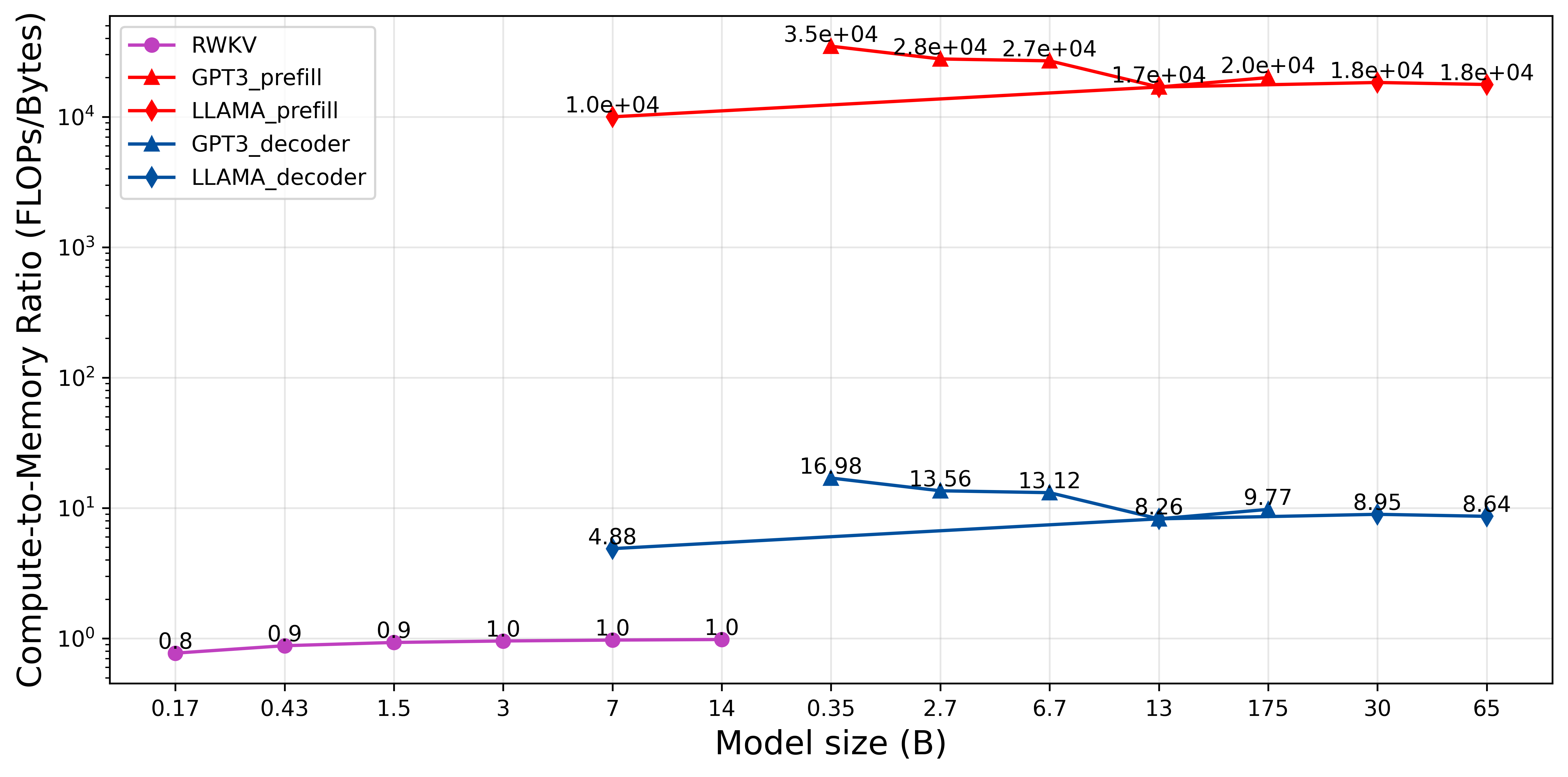}
    \caption{Compute-to-memory-ratio for different models.}
    \label{compute_to_memory_ratio}
\end{figure}

The figure \ref{compute_to_memory_ratio} compares the Compute-to-Memory Ratio (FLOPs per byte) across various models and highlights that RWKV consistently exhibits the lowest ratio, indicating that its operations rely more on memory access rather than intensive computations, compared to models like GPT-3 and LLAMA. This characteristic makes RWKV particularly well-suited for acceleration through weight quantization, as its lower computational demands relative to memory usage allow for more significant gains in inference speed, especially when optimizing for resource-constrained environments.

\input{tabs/appendix_visual_res}

\subsection{Additional Results}\label{appendix c}
% \begin{figure}[H]
% \vspace{-2mm}
%     \centering
%     \includegraphics[width=0.4\textwidth]{imgs/percent.png}
%     \caption{Comparison of SQ proportion between RWKV and LLaMA Models.}
%     \label{SQ_percent}
% \vspace{-5mm}
% \end{figure}

% Table~\ref{tab_kmeans_loss} presents the average relative error of weight quantization using K-means clustering for the RWKV family and the LLaMA family separately. Under the settings of $\tau_c=1.5$ and $\tau_f=50$, Figure~\ref{SQ_percent} shows that approximately 60\% of the layers in the RWKV family are categorized as suitable for uniform quantization, whereas the proportion is only about 10\% for the LLaMA family. This further demonstrates that the RWKV models have a significantly higher number of uniform layers.

We provide a comprehensive presentation of our results across various datasets to complement the main paper. Specifically, the results include:
    \begin{enumerate}[label=\textbullet]
        \item Complete comparison of the results under different quantization settings for Vision RWKV models.(Table \ref{appendix_VisualResults}).
        \item Complete comparison of the perplexity score on LAMBADA and averaged accuracy on zero-shot common sense reasoning tasks on RWKV7 (Tab \ref{tab:appendix_main_RWKV7}) and RWKV6 (Table \ref{tab:appendix_main_RWKV6}). 
        \item Validate the effectiveness of codebook optimization for element-wise multiplication. (Table \ref{tab:appendix_main_mul}). 
    \end{enumerate}

\input{tabs/appdix_quantitative_results}
\input{tabs/appendix_ablition_mul}
% \vspace{10 mm}
\subsection{Limitations and Future Work}\label{appendix d}
Our proposed RWKVQuant framework relies on the coarse-to-fine proxy introduced in Section~\ref{subsec_hybrid}, where 
$\tau_c$ and $\tau_f$ play a important role in determining the appropriate quantization method for each layer. In our experiments, these values were empirically set based on different model configurations, ensuring that the 3.25 bpw SQ proportion is approximately one-tenth and nine-tenths for 3.5 bpw VQ across different models. However, this allocation might not reflect the most balanced or effective proportion.

\input{tabs/different_muf}

As shown in Table~\ref{different_result}, we conducted multiple comparative experiments on the RWKV language model with varying $\tau_c$ and $\tau_f$. The results demonstrate that when $\tau_c$ is larger than the optimal value, the final accuracy approaches that of directly using uniform quantization, whereas a smaller $\tau_c$ yields results closer to codebook quantization. Moreover, with an appropriately chosen $\tau_c$, the setting of $\tau_f$ has a direct impact on the final accuracy.Therefore, in future work, we will further explore how to determine appropriate values for $\tau_c$ and $\tau_f$:
    \begin{enumerate}[label=\textbullet]
        \item Based on proper initialization, we plan to utilize fine-tuning to achieve the optimal configuration for different models.
        \item  At the same time, we will remove the fixed constraint of bpw being 3.275 and, when selecting $\tau_c$ and $\tau_f$, consider the trade-off between compression rate and post-quantization model performance to meet different accuracy requirements across various scenarios.
        \item In addition, we plan to use the proposed coarse-to-fine proxy to determine whether specific features at a finer granularity are better suited for SQ or VQ quantization, such as channel-level or block-level finer granularity.
    \end{enumerate}

%% file: tabs/appendix_visual_res.tex
\begin{table*}[h]
\renewcommand\arraystretch{1.4} % 调整行高
\centering
\vspace{1mm}
\caption{Complete comparative results under different quantization settings for Vision RWKV models. For classification tasks, we report the Top-1 Accuracy on ImageNet. For detection tasks, the Box AP is evaluated on Coco, while for segmentation tasks, the mIoU is measured on ADE20K.}
\vspace{1mm}
\label{appendix_VisualResults}
\setlength{\tabcolsep}{0.5mm} % 调整列间距
\resizebox{\textwidth}{!}{%
\begin{tabular}{c|c|ccc|ccc|ccc|ccc|ccc|ccc|ccc}
\hline
\multirow{2}{*}{\textbf{Bpw.}} & \multirow{2}{*}{\textbf{Method}} & \multicolumn{3}{c|}{\textbf{RWKV-T}} & \multicolumn{3}{c|}{\textbf{RWKV-S}} & \multicolumn{3}{c|}{\textbf{RWKV-B}} & \multicolumn{3}{c|}{\textbf{RWKV-L}} & \multicolumn{3}{c|}{\textbf{RWKV6-T}} & \multicolumn{3}{c|}{\textbf{RWKV-L}} & \multicolumn{3}{c}{\textbf{RWKV6-T}}\\
\noalign{\vspace{0.1em}}\cdashline{3-23}\noalign{\vspace{0.1em}}
             &                 & \textbf{Cls.} & \textbf{Det.} & \textbf{Seg.} & \textbf{Cls.} & \textbf{Det.} & \textbf{Seg.} 
             & \textbf{Cls.} & \textbf{Det.} & \textbf{Seg.}
             & \textbf{Cls.} & \textbf{Det.} & \textbf{Seg.}
             & \textbf{Cls.} & \textbf{Det.} & \textbf{Seg.}
             & \textbf{Cls.} & \textbf{Det.} & \textbf{Seg.}
             & \textbf{Cls.} & \textbf{Det.} & \textbf{Seg.}
             \\
\hline
16       & FloatingPoint      & 75.10 & 41.70 & 43.30 & 80.10 & 44.80 & 47.20 
                              & 82.00  & 46.80  & 49.20
                              & -     & 50.6  & 53.50
                              & 76.60  & - & -
                              & 81.10  & - & -
                              & 82.60  & - & - 
                                \\
\hline
\multirow{4}{*}{3.5} 
             & GPTQ           & 69.74  & 39.85 & 41.20 & 78.30 & 43.37 & 45.50 
                              & 81.42  & 46.14  & 48.64
                              & -      & 50.30   & 53.26
                              & 72.79  & - & -
                              & 80.13  & - & -
                              & 82.31  & - & -             
                                \\
             & AWQ            & 68.50  & 39.03 & 38.88 & 78.00 & 42.90 & 42.88 
                              & 81.15  & 45.70  & 48.55
                              & -      & 50.19 & 53.18
                              & 71.46  & - & -
                              & 79.74  & - & -
                              & 82.09  & - & -     
                                \\          
             & GPTVQ          & 70.61  & 40.14 & 41.65 & 78.65 & 44.03 & 45.00 
                              & 81.37  & 46.23  & 48.70
                              & -      & 50.28   & 52.90
                              & 73.22  & - & -
                              & 80.23  & - & -
                              & 82.25  & - & -                               
                                \\
             & VPTQ           & 67.21  & 39.02  & 40.14 & 76.40 & 42.01 & 43.54 
                              & 80.29  & 45.02  & 48.68
                              & -      & 49.10  & 51.45
                              & 70.36  & - & -
                              & 77.75  & - & -
                              & 81.31  & - & -                
                                \\
\hline
3.275        & \textbf{Ours} & 70.41  & 40.22 & 41.70 &78.74 & 43.85 & 46.09 
                              & 81.58  & 46.40  & 48.49
                              & -      & 50.31   & 52.95
                              & 73.13  & - & -
                              & 80.24  & - & -
                              & 82.32  & - & -    
\\
\hline
\end{tabular}%
}
\vspace{1mm}
\end{table*}

%% file: tabs/appdix_quantitative_results.tex
\begin{table}[H]
\renewcommand\arraystretch{0.8}
\centering
\caption{\small Complete comparison of perplexity on LAMBADA and averaged accuracy on nine Zero-Shot tasks. For all methods except ours and floating-point, we report metrics under both bpw settings of 3.5 and 3.25.}
\vspace{-6em}
\label{tab:appendix_main_RWKV7}
\setlength{\tabcolsep}{1mm}
{\resizebox{\textwidth}{!}{
\begin{tabular}{c|c|l|cccccccccc|c}
& & & & & & & & & & & &\\
& & & & & & & & & & & &\\
& & & & & & & & & & & &\\
& & & & & & & & & & & &\\
& & & & & & & & & & & &\\
& & & & & & & & & & & &\\
\noalign{\vspace{0.2em}}\hline\noalign{\vspace{0.1em}}
\noalign{\vspace{0.1em}}\hline\noalign{\vspace{0.2em}}
\multirow{2}{*}{\textbf{Model}} & \multirow{2}{*}{\textbf{Bpw.}} & \multirow{2}{*}{\textbf{Method}} & \textbf{ARC-c} & \textbf{ARC-e
} & \textbf{HQA.} & \textbf{HellaS.
} & \textbf{Lam.
} & \textbf{OBQA} & \textbf{PIQA
} & \textbf{SCIQ
} & \textbf{WinoG.
}  & \textbf{Avg.
}  & \textbf{Wiki2} \\ 
&  & & ($\uparrow$) & ($\uparrow$) & ($\uparrow$) & ($\uparrow$) & ($\uparrow$) & ($\uparrow$) & ($\uparrow$) & ($\uparrow$) & ($\uparrow$) & ($\uparrow$) & ($\downarrow$) \\
\noalign{\vspace{0.2em}}\hline\noalign{\vspace{0.2em}}
\multirow{22}{*}{RWKV-7 0.1B} & 16 & Full Precision & 19.7 & 47.90 & 25.12 & 31.59 & 45.62 & 17.2 & 65.61 & 81.8 & 52.6 & 43.02 & 14.21  \\
\noalign{\vspace{0.2em}}\cdashline{2-14}\noalign{\vspace{0.2em}}
         & \multirow{8}[0]{*}{3.25} & RTN   & 19.53 & 40.66 & 23.70 & 30.12 & 12.83 & 14.4 & 62.73 & 70.2 & 51.93 & 36.22 & 152.82  \\
          &       & GPTQ   & 20.39  & 42.17  & 23.52  & 30.59  & 20.47  & 15.40   & 62.62  & 73.60   & 52.50   & 37.92  & 63.54   \\
          &       & AWQ    & 20.13  & 41.16  & 22.72  & 29.88  & 15.20  & 25.44   & 61.15  & 76.70   & 52.40   & 38.31  & 132.06  \\
          &       & QuaRot & 19.96  & 39.94  & 22.46  & 29.38  & 12.28  & 13.80   & 60.93  & 58.80   & 53.19   & 34.53  & 243.99  \\
          &       & KMeans & 18.94  & 43.56  & 23.81  & 29.85  & 19.93  & 15.00   & 63.49  & 75.70   & 53.59   & 38.21  & 87.06  \\
          &       & GPTVQ  & 18.94  & 43.93  & 22.75  & 30.04  & 38.06  & 16.40   & 63.36  & 77.00   & 51.77   & 40.25  & 23.75  \\
          &       & VPTQ   & 18.94  & 39.60  & 21.84  & 28.59  & 17.50  & 13.20   & 59.90  & 71.90   & 50.51   & 35.78  & 128.59  \\

\noalign{\vspace{0.2em}}\cdashline{2-14}\noalign{\vspace{0.2em}}

          & \multirow{7}[0]{*}{3.5} & RTN   & 19.36 & 43.47 & 24.50 & 30.17 & 18.13 & 16.20 & 64.09 & 74.40 & 52.48 & 38.08 & 81.14 \\
          &       & GPTQ   & 18.68 & 44.06 & 24.14 & 30.69 & 26.74 & 17.80 & 63.76 & 74.30 & 53.82 & 39.33 & 40.16 \\
          &       & AWQ    & 19.88 & 40.40 & 22.97 & 29.52 & 14.92 & 14.00 & 59.99 & 71.50 & 52.64 & 36.20 & 55.72 \\
          &       & QuaRot & 19.88 & 44.40 & 22.72 & 30.16 & 15.36 & 15.60 & 62.78 & 72.40 & 52.09 & 37.26 & 126.19   \\
          &       & KMeans & 19.19 & 44.02 & 24.14 & 29.98 & 29.44 & 15.00 & 63.54 & 77.40 & 53.27 & 39.55 & 36.23  \\
          &       & GPTVQ  & 19.11 & 42.88 & 22.68 & 30.25 & 38.35 & 15.80 & 63.11 & 74.40 & 54.30 & 40.10 & 25.82 \\
          &       & VPTQ   & 18.51 & 40.74 & 22.42 & 28.82 & 23.17 & 13.80 & 61.31 & 72.90 & 51.93 & 37.07 & 74.70 \\
          
\noalign{\vspace{0.2em}}\cdashline{2-14}\noalign{\vspace{0.2em}}
          & \multirow{1}[0]{*}{3.275}      & \cellcolor[rgb]{ .906,  .902,  .902}\textbf{Ours} & \cellcolor[rgb]{ .906,  .902,  .902}19.73 
          & \cellcolor[rgb]{ .906,  .902,  .902}42.17  
          & \cellcolor[rgb]{ .906,  .902,  .902}24.14  
          & \cellcolor[rgb]{ .906,  .902,  .902}30.02  
          & \cellcolor[rgb]{ .906,  .902,  .902}42.25  
          & \cellcolor[rgb]{ .906,  .902,  .902}17.00  
          & \cellcolor[rgb]{ .906,  .902,  .902}63.10  
          & \cellcolor[rgb]{ .906,  .902,  .902}79.10  
          & \cellcolor[rgb]{ .906,  .902,  .902}52.30 
          & \cellcolor[rgb]{ .906,  .902,  .902}41.10  
          & \cellcolor[rgb]{ .906,  .902,  .902}18.41  \\
\noalign{\vspace{0.2em}}\hline\noalign{\vspace{0.2em}}

\multirow{22}{*}{RWKV-7 0.5B} & 16 & Full Precision & 23.80 & 57.11 & 28.55 & 38.28 & 57.85 & 20.80 & 69.31 & 86.6 & 55.72 & 48.67 & 7.21  \\
\noalign{\vspace{0.2em}}\cdashline{2-14}\noalign{\vspace{0.2em}}
         & \multirow{8}[0]{*}{3.25} & RTN   & 21.84 & 45.28 & 23.66 & 32.64 & 26.45 & 19.20 & 64.96 & 73.4 & 52.56 & 39.99 & 57.11  \\
          &       & GPTQ   & 22.61 & 52.77 & 26.14 & 35.92 & 42.52 & 20.80 & 67.46 & 79.60 & 55.40 & 43.69 & 15.97   \\
          &       & AWQ    & 19.62 & 43.09 & 22.93 & 32.64 & 39.34 & 17.00 & 62.57 & 79.20 & 54.14 & 41.16 & 23.29  \\
          &       & QuaRot & 22.18 & 48.69 & 24.76 & 33.85 & 20.22 & 19.00 & 65.01 & 73.80 & 54.06 & 40.17 & 76.89  \\
          &       & KMeans & 23.46 & 52.48 & 26.29 & 36.19 & 39.41 & 20.00 & 67.35 & 81.50 & 54.69 & 44.59 & 20.19  \\
          &       & GPTVQ  & 21.50 & 51.30 & 26.14 & 34.80 & 46.19 & 19.40 & 66.64 & 82.40 & 54.38 & 43.64 & 14.15  \\
          &       & VPTQ   & 18.68 & 42.97 & 22.86 & 31.15 & 34.81 & 16.40 & 62.73 & 78.80 & 52.88 & 40.14 & 30.63  \\

\noalign{\vspace{0.2em}}\cdashline{2-14}\noalign{\vspace{0.2em}}

          & \multirow{7}[0]{*}{3.5} & RTN   & 18.13 & 19.36 & 43.47 & 24.50 & 30.17 & 16.20 & 64.09 & 74.40 & 52.48 & 38.08 & 81.14 \\
          &       & GPTQ   & 22.26 & 53.11 & 25.41 & 36.51 & 45.78 & 21.40 & 67.51 & 83.10 & 56.51 & 45.73 & 13.07 \\
          &       & AWQ    & 21.33 & 43.60 & 22.57 & 33.06 & 45.14 & 18.80 & 61.75 & 82.60 & 52.80 & 42.40 & 16.98 \\
          &       & QuaRot & 28.33 & 20.13 & 46.38 & 24.54 & 32.30 & 18.80 & 65.23 & 80.40 & 51.46 & 40.84 & 40.38   \\
          &       & KMeans & 20.73 & 47.55 & 25.05 & 33.58 & 41.39 & 17.40 & 65.83 & 82.00 & 54.14 & 43.07 & 17.05  \\
          &       & GPTVQ  & 21.92 & 52.18 & 25.38 & 34.90 & 52.05 & 17.80 & 67.51 & 81.60 & 53.82 & 44.13 & 10.88  \\
          &       & VPTQ   & 19.79 & 44.73 & 23.12 & 31.62 & 37.78 & 16.40 & 63.98 & 78.90 & 53.27 & 41.06 & 25.03 \\
          
\noalign{\vspace{0.2em}}\cdashline{2-14}\noalign{\vspace{0.2em}}
          & \multirow{1}[0]{*}{3.275}      & \cellcolor[rgb]{ .906,  .902,  .902}\textbf{Ours} 
          & \cellcolor[rgb]{ .906,  .902,  .902}22.61  
          & \cellcolor[rgb]{ .906,  .902,  .902}53.11  
          & \cellcolor[rgb]{ .906,  .902,  .902}25.41  
          & \cellcolor[rgb]{ .906,  .902,  .902}34.45  
          & \cellcolor[rgb]{ .906,  .902,  .902}53.39  
          & \cellcolor[rgb]{ .906,  .902,  .902}19.60 
          & \cellcolor[rgb]{ .906,  .902,  .902}66.53  
          & \cellcolor[rgb]{ .906,  .902,  .902}84.20
          & \cellcolor[rgb]{ .906,  .902,  .902}54.80  
          & \cellcolor[rgb]{ .906,  .902,  .902}46.01
          & \cellcolor[rgb]{ .906,  .902,  .902}9.39 \\
\noalign{\vspace{0.2em}}\hline\noalign{\vspace{0.2em}}

    \multirow{22}{*}{RWKV-7 1.47B} & 16 & Full Precision & 31.65 & 65.40 & 32.12 & 46.55 & 66.97 & 26.00 & 72.68 & 90.00 & 64.40 & 55.08 & 4.84  \\
\noalign{\vspace{0.2em}}\cdashline{2-14}\noalign{\vspace{0.2em}}
         & \multirow{8}[0]{*}{3.25} & RTN   & 21.69 & 53.44 & 25.65 & 35.81 & 46.63 & 17.90 & 65.39 & 85.14 & 57.55 & 45.46 & 11.43  \\
          &       & GPTQ   & 29.01 & 59.93 & 28.99 & 42.74 & 55.52 & 23.60 & 70.34 & 87.30 & 62.98 & 51.15 & 7.93   \\
          &       & AWQ    & 22.35 & 44.91 & 22.72 & 35.22 & 45.04 & 16.00 & 61.26 & 85.50 & 59.58 & 43.62 & 15.27  \\
          &       & QuaRot & 29.43 & 63.04 & 30.01 & 42.24 & 52.30 & 22.80 & 70.78 & 88.10 & 58.64 & 50.81 & 9.39  \\
          &       & KMeans & 29.86 & 60.39 & 28.92 & 43.07 & 57.75 & 23.20 & 70.78 & 90.20 & 62.90 & 51.89 & 7.04  \\
          &       & GPTVQ  & 27.21 & 61.48 & 28.51 & 42.95 & 64.84 & 22.80 & 70.34 & 88.40 & 62.03 & 52.06 & 5.54  \\
          &       & VPTQ   & 22.78 & 51.55 & 25.23 & 36.13 & 49.31 & 18.20 & 66.21 & 85.40 & 57.69 & 45.83 & 11.13  \\

\noalign{\vspace{0.2em}}\cdashline{2-14}\noalign{\vspace{0.2em}}

          & \multirow{7}[0]{*}{3.5} & RTN   & 29.94 & 61.82 & 30.45 & 43.81 & 54.84 & 24.2 & 71.27 & 88.1 & 61.32 & 51.74 & 7.89 \\
          &       & GPTQ   & 30.54 & 61.65 & 29.79 & 43.96 & 59.23 & 23.80 & 71.65 & 88.90 & 61.06 & 52.28 & 6.55 \\
          &       & AWQ    & 22.44 & 46.00 & 21.95 & 35.30 & 52.65 & 17.20 & 61.86 & 84.50 & 59.66 & 44.61 & 10.71 \\
          &       & QuaRot & 30.80 & 62.28 & 30.85 & 43.27 & 55.64 & 23.20 & 71.87 & 89.20 & 60.77 & 51.98 & 7.94   \\
          &       & KMeans & 29.69 & 61.44 & 30.48 & 43.92 & 61.58 & 24.60 & 71.32 & 89.90 & 62.03 & 52.77 & 6.57  \\
          &       & GPTVQ  & 28.32 & 61.57 & 29.24 & 42.63 & 64.18 & 21.20 & 70.83 & 88.60 & 62.66 & 52.13 & 5.51  \\
          &       & VPTQ   & 23.54 & 54.25 & 26.44 & 37.70 & 52.24 & 19.04 & 67.73 & 86.90 & 58.64 & 47.38 & 9.52 \\
          
\noalign{\vspace{0.2em}}\cdashline{2-14}\noalign{\vspace{0.2em}}
          & \multirow{1}[0]{*}{3.275}      & \cellcolor[rgb]{ .906,  .902,  .902}\textbf{Ours} 
          & \cellcolor[rgb]{ .906,  .902,  .902}28.15  
          & \cellcolor[rgb]{ .906,  .902,  .902}61.32  
          & \cellcolor[rgb]{ .906,  .902,  .902}29.35  
          & \cellcolor[rgb]{ .906,  .902,  .902}43.24  
          & \cellcolor[rgb]{ .906,  .902,  .902}65.38  
          & \cellcolor[rgb]{ .906,  .902,  .902}22.60
          & \cellcolor[rgb]{ .906,  .902,  .902}71.70  
          & \cellcolor[rgb]{ .906,  .902,  .902}88.90
          & \cellcolor[rgb]{ .906,  .902,  .902}61.01  
          & \cellcolor[rgb]{ .906,  .902,  .902}52.40
          & \cellcolor[rgb]{ .906,  .902,  .902}5.24 \\
\noalign{\vspace{0.2em}}\hline\noalign{\vspace{0.2em}}
\hline\noalign{\vspace{0.2em}}
\end{tabular}}}
\end{table}

\begin{table}[H]
\renewcommand\arraystretch{0.7}
\centering
\caption{\small Complete comparison of perplexity on LAMBADA and averaged accuracy on nine Zero-Shot tasks. For all methods except ours and floating-point, we report metrics under both bpw settings of 3.5 and 3.25.}
\vspace{-6em}
\label{tab:appendix_main_RWKV6}
\setlength{\tabcolsep}{1mm}
{\resizebox{\textwidth}{!}{
\begin{tabular}{c|c|l|cccccccccc|c}
& & & & & & & & & & & &\\
& & & & & & & & & & & &\\
& & & & & & & & & & & &\\
& & & & & & & & & & & &\\
& & & & & & & & & & & &\\
& & & & & & & & & & & &\\
& & & & & & & & & & & &\\
& & & & & & & & & & & &\\
\noalign{\vspace{0.2em}}\hline\noalign{\vspace{0.1em}}
\noalign{\vspace{0.1em}}\hline\noalign{\vspace{0.2em}}
\multirow{2}{*}{\textbf{Model}} & \multirow{2}{*}{\textbf{Bpw.}} & \multirow{2}{*}{\textbf{Method}} & \textbf{ARC-c} & \textbf{ARC-e
} & \textbf{HQA.} & \textbf{HellaS.
} & \textbf{Lam.
} & \textbf{OBQA} & \textbf{PIQA
} & \textbf{SCIQ
} & \textbf{WinoG.
}  & \textbf{Avg.
}  & \textbf{Wiki2} \\ 
&  & & ($\uparrow$) & ($\uparrow$) & ($\uparrow$) & ($\uparrow$) & ($\uparrow$) & ($\uparrow$) & ($\uparrow$) & ($\uparrow$) & ($\uparrow$) & ($\uparrow$) & ($\downarrow$) \\
\noalign{\vspace{0.2em}}\hline\noalign{\vspace{0.2em}}
\multirow{22}{*}{RWKV-6 1B} & 16 & Full Precision & 31.22 & 64.40 & 30.45 & 46.34 & 67.11 & 25.20 & 74.26 & 89.60 & 60.93 & 54.39 & 4.60  \\
\noalign{\vspace{0.2em}}\cdashline{2-14}\noalign{\vspace{0.2em}}
         & \multirow{8}[0]{*}{3.25} & RTN   & 25.25 & 55.43 & 27.13 & 42.59 & 59.73 & 24.00 & 71.00 & 84.80 & 58.64 & 49.84 & 6.39  \\
          &       & GPTQ   & 25.17 & 58.62 & 27.06 & 42.07 & 60.18 & 24.00 & 71.16 & 86.70 & 60.06 & 50.55 & 6.43   \\
          &       & AWQ    & 23.72 & 50.16 & 23.12 & 37.34 & 40.03 & 19.60 & 65.12 & 83.50 & 57.14 & 44.41 & 17.97  \\
          &       & QuaRot & 27.30 & 51.89 & 26.65 & 37.42 & 32.08 & 21.20 & 67.84 & 85.00 & 55.80 & 45.02 & 29.38  \\
          &       & KMeans & 27.98 & 55.21 & 26.80 & 40.03 & 41.24 & 20.20 & 69.53 & 85.20 & 57.06 & 47.02 & 15.93  \\
          &       & GPTVQ  & 24.65 & 57.82 & 27.20 & 39.67 & 62.33 & 22.00 & 70.02 & 86.30 & 58.80 & 49.86 & 6.11  \\
          &       & VPTQ   & 21.33 & 49.45 & 23.59 & 34.65 & 42.95 & 17.80 & 65.56 & 84.40 & 55.32 & 43.89 & 14.67  \\

\noalign{\vspace{0.2em}}\cdashline{2-14}\noalign{\vspace{0.2em}}

          & \multirow{7}[0]{*}{3.5} & RTN & 29.43 & 60.81 & 27.46 & 43.64 & 61.11 & 22.60 & 72.52 & 88.40 & 59.90 & 51.76 & 5.83 \\
          &       & GPTQ   & 26.53 & 59.55 & 27.53 & 43.63 & 61.61 & 23 & 71.76 & 87.9 & 59.58 & 51.23 & 5.86 \\
          &       & AWQ    & 24.57 & 53.07 & 22.90 & 38.57 & 50.48 & 20.40 & 66.43 & 85.20 & 58.87 & 51.20 & 10.97 \\
          &       & QuaRot & 26.19 & 54.67 & 26.91 & 39.21 & 41.34 & 23.60 & 68.60 & 85.70 & 56.82 & 47.00 & 16.29   \\
          &       & KMeans & 28.07 & 60.26 & 28.11 & 41.84 & 51.50 & 22.40 & 70.23 & 87.80 & 60.14 & 50.03 & 8.24  \\
          &       & GPTVQ  & 27.81 & 58.71 & 26.18 & 40.41 & 61.96 & 22.20 & 70.83 & 87.30 & 57.22 & 50.29 & 5.74 \\
          &       & VPTQ   & 20.87 & 49.99 & 23.41 & 34.37 & 43.41 & 17.80 & 64.20 & 85.50 & 54.85 & 43.82 & 14.74 \\
          
\noalign{\vspace{0.2em}}\cdashline{2-14}\noalign{\vspace{0.2em}}
          & \multirow{1}[0]{*}{3.275}      & \cellcolor[rgb]{ .906,  .902,  .902}\textbf{Ours} & \cellcolor[rgb]{ .906,  .902,  .902}27.90
          & \cellcolor[rgb]{ .906,  .902,  .902}60.35  
          & \cellcolor[rgb]{ .906,  .902,  .902}28.08  
          & \cellcolor[rgb]{ .906,  .902,  .902}42.51  
          & \cellcolor[rgb]{ .906,  .902,  .902}63.87  
          & \cellcolor[rgb]{ .906,  .902,  .902}23.00  
          & \cellcolor[rgb]{ .906,  .902,  .902}71.38  
          & \cellcolor[rgb]{ .906,  .902,  .902}87.90  
          & \cellcolor[rgb]{ .906,  .902,  .902}60.22 
          & \cellcolor[rgb]{ .906,  .902,  .902}51.69  
          & \cellcolor[rgb]{ .906,  .902,  .902}5.29  \\
\noalign{\vspace{0.2em}}\hline\noalign{\vspace{0.2em}}

\multirow{22}{*}{RWKV-6 3B} & 16 & Full Precision & 35.58 & 71.33 & 33.29 & 50.53 & 71.32 & 28.00 & 76.15 & 92.30 & 66.45 & 58.32 & 3.83  \\
\noalign{\vspace{0.2em}}\cdashline{2-14}\noalign{\vspace{0.2em}}
         & \multirow{8}[0]{*}{3.25} & RTN   & 30.54 & 64.52 & 28.55 & 47.21 & 66.87 & 24.8 & 73.34 & 88.8 & 62.98 & 54.17 & 4.71  \\
          &       & GPTQ   & 31.14 & 65.06 & 28.77 & 46.92 & 65.65 & 25.40 & 73.83 & 85.1 & 63.61 & 53.94 & 4.88   \\
          &       & AWQ    & 26.02 & 55.13 & 23.48 & 40.42 & 47.53 & 17.00 & 66.05 & 87.80 & 61.79 & 47.24 & 11.97  \\
          &       & QuaRot & 29.43 & 57.82 & 28.55 & 41.59 & 38.35 & 24.00 & 69.74 & 83.80 & 60.93 & 48.24 & 22.67  \\
          &       & KMeans & 31.99 & 66.41 & 30.99 & 44.62 & 52.26 & 25.60 & 72.41 & 92.10 & 61.24 & 53.06 & 8.27  \\
          &       & GPTVQ  & 29.77 & 64.52 & 28.84 & 44.56 & 69.92 & 24.40 & 72.57 & 91.20 & 62.35 & 54.23 & 4.31  \\
          &       & VPTQ   & 25.00 & 56.56 & 25.20 & 37.93 & 55.95 & 18.40 & 67.57 & 86.20 & 61.24 & 48.22 & 7.77  \\

\noalign{\vspace{0.2em}}\cdashline{2-14}\noalign{\vspace{0.2em}}

          & \multirow{7}[0]{*}{3.5} & RTN   & 31.56 & 66.75 & 30.78 & 47.91 & 68.02 & 26.00 & 74.75 & 90.50 & 53.53 & 54.42 & 4.50 \\
          &       & GPTQ   & 31.48 & 67.17 & 29.80 & 47.65 & 67.49 & 25.80 & 74.26 & 90.30 & 63.22 & 55.24 & 4.54 \\
          &       & AWQ    & 28.41 & 54.67 & 24.03 & 39.96 & 53.06 & 19.20 & 66.15 & 88.20 & 61.95 & 48.40 & 8.77 \\
          &       & QuaRot & 31.22 & 61.48 & 29.46 & 43.04 & 42.99 & 29.20 & 71.21 & 89.10 & 61.48 & 51.01 & 16.99   \\
          &       & KMeans & 34.72 & 67.46 & 31.43 & 46.63 & 58.53 & 25.60 & 73.77 & 91.90 & 64.95 & 54.99 & 6.28  \\
          &       & GPTVQ  & 30.04 & 65.61 & 29.57 & 45.39 & 69.84 & 24.60 & 73.12 & 89.90 & 65.58 & 54.85 & 4.12  \\
          &       & VPTQ   & 25.59 & 56.81 & 25.52 & 38.09 & 53.72 & 19.60 & 68.49 & 87.70 & 60.69 & 48.46 & 8.62 \\
          
\noalign{\vspace{0.2em}}\cdashline{2-14}\noalign{\vspace{0.2em}}
          & \multirow{1}[0]{*}{3.275}      & \cellcolor[rgb]{ .906,  .902,  .902}\textbf{Ours} 
          & \cellcolor[rgb]{ .906,  .902,  .902}30.97 
          & \cellcolor[rgb]{ .906,  .902,  .902}65.74  
          & \cellcolor[rgb]{ .906,  .902,  .902}30.12  
          & \cellcolor[rgb]{ .906,  .902,  .902}46.90  
          & \cellcolor[rgb]{ .906,  .902,  .902}71.18  
          & \cellcolor[rgb]{ .906,  .902,  .902}25.00 
          & \cellcolor[rgb]{ .906,  .902,  .902}74.53  
          & \cellcolor[rgb]{ .906,  .902,  .902}92.20
          & \cellcolor[rgb]{ .906,  .902,  .902}65.51  
          & \cellcolor[rgb]{ .906,  .902,  .902}55.79
          & \cellcolor[rgb]{ .906,  .902,  .902}3.88 \\
\noalign{\vspace{0.2em}}\hline\noalign{\vspace{0.2em}}

\multirow{22}{*}{RWKV-6 7B} & 16 & Full Precision & 41.70 & 75.25 & 35.66 & 55.82 & 75.35 & 31.4 & 78.18 & 93.80 & 68.11 & 61.69 & 3.21  \\
\noalign{\vspace{0.2em}}\cdashline{2-14}\noalign{\vspace{0.2em}}
         & \multirow{8}[0]{*}{3.25} & RTN   & 35.66 & 70.32 & 32.53 & 51.85 & 70.04 & 28.20 & 77.20 & 92.60 & 66.69 & 58.34 & 3.87  \\
          &       & GPTQ   & 38.56 & 72.51 & 31.69 & 51.98 & 71.78 & 29.40 & 77.14 & 93.50 & 67.00 & 59.28 & 3.72   \\
          &       & AWQ    & 29.01 & 57.40 & 23.92 & 40.72 & 52.44 & 20.60 & 67.3  & 88.10 & 65.91 & 49.48 & 8.33  \\
          &       & QuaRot & 32.25 & 67.46 & 31.36 & 47.39 & 51.43 & 28.40 & 74.59 & 89.60 & 66.21 & 54.29 & 8.81  \\
          &       & KMeans & 38.73 & 72.72 & 34.71 & 51.41 & 61.85 & 29.00 & 76.93 & 81.50 & 54.69 & 55.72 & 4.69  \\
          &       & GPTVQ  & 36.34 & 71.63 & 32.09 & 49.78 & 74.17 & 28.40 & 74.64 & 92.40 & 67.71 & 58.57 & 3.49  \\
          &       & VPTQ   & 28.83 & 63.80 & 28.18 & 42.15 & 66.23 & 21.20 & 71.32 & 91.10 & 64.79 & 53.06 & 4.75  \\

\noalign{\vspace{0.2em}}\cdashline{2-14}\noalign{\vspace{0.2em}}

          & \multirow{7}[0]{*}{3.5} & RTN   & 37.37 & 70.79 & 32.53 & 53.26 & 73.06 & 29.80 & 76.93 & 90.5 & 68.58 & 59.20 & 3.59 \\
          &       & GPTQ   & 39.33 & 73.40 & 32.96 & 52.95 & 71.96 & 30.60 & 77.2  & 94.20 & 68.11 & 60.07 & 3.68 \\
          &       & AWQ    & 29.01 & 57.40 & 23.92 & 40.72 & 61.54 & 20.60 & 67.3  & 88.10 & 65.91 & 50.50 & 7.07 \\
          &       & QuaRot & 35.32 & 67.76 & 31.36 & 49.68 & 60.18 & 31.40 & 75.46 & 94.50 & 66.92 & 56.95 & 6.44   \\
          &       & KMeans & 39.84 & 74.03 & 34.35 & 53.01 & 67.48 & 28.60 & 76.98 & 93.90 & 67.48 & 59.51 & 3.96  \\
          &       & GPTVQ  & 38.50 & 72.50 & 31.69 & 51.98 & 75.04 & 29.40 & 77.71 & 93.50 & 67.00 & 59.70 & 3.30  \\
          &       & VPTQ   & 28.78 & 62.33 & 27.83 & 42.45 & 67.86 & 21.00 & 71.49 & 90.40 & 64.48 & 52.95 & 4.47 \\
          
\noalign{\vspace{0.2em}}\cdashline{2-14}\noalign{\vspace{0.2em}}
          & \multirow{1}[0]{*}{3.275}      & \cellcolor[rgb]{ .906,  .902,  .902}\textbf{Ours} 
          & \cellcolor[rgb]{ .906,  .902,  .902}37.62  
          & \cellcolor[rgb]{ .906,  .902,  .902}72.93  
          & \cellcolor[rgb]{ .906,  .902,  .902}33.63  
          & \cellcolor[rgb]{ .906,  .902,  .902}52.03  
          & \cellcolor[rgb]{ .906,  .902,  .902}75.70 
          & \cellcolor[rgb]{ .906,  .902,  .902}30.40
          & \cellcolor[rgb]{ .906,  .902,  .902}76.60  
          & \cellcolor[rgb]{ .906,  .902,  .902}95.10
          & \cellcolor[rgb]{ .906,  .902,  .902}67.71  
          & \cellcolor[rgb]{ .906,  .902,  .902}60.19
          & \cellcolor[rgb]{ .906,  .902,  .902}3.23 \\
\noalign{\vspace{0.2em}}\hline\noalign{\vspace{0.2em}}

    \multirow{22}{*}{RWKV-6 14B} & 16 & Full Precision & 44.70 & 76.97 & 37.05 & 58.76 & 76.23 & 33.6 & 79.59 & 94.7 & 71.27 & 63.65 & 3.02  \\
\noalign{\vspace{0.2em}}\cdashline{2-14}\noalign{\vspace{0.2em}}
         & \multirow{8}[0]{*}{3.25} & RTN   & 39.76 & 73.98 & 32.93 & 55.66 & 74.03 & 31.80 & 79.10 & 93.00 & 70.24 & 61.16 & 3.34  \\
          &       & GPTQ   & 38.22 & 69.99 & 33.36 & 55.28 & 74.00 & 31.40 & 78.18 & 91.50 & 69.69 & 60.18 & 3.43   \\
          &       & AWQ    & 28.15 & 55.34 & 23.77 & 38.82 & 58.59 & 20.60 & 65.61 & 88.90 & 64.40 & 49.35 & 8.18  \\
          &       & QuaRot & 36.51 & 69.27 & 32.78 & 50.38 & 40.09 & 28.20 & 76.06 & 91.60 & 67.95 & 54.76 & 14.05 \\
          &       & KMeans & 43.85 & 74.16 & 35.55 & 54.84 & 67.81 & 32.40 & 78.18 & 94.80 & 71.27 & 61.40 & 4.61  \\
          &       & GPTVQ  & 35.65 & 71.66 & 30.45 & 52.33 & 76.46 & 30.14 & 76.31 & 93.24 & 70.44 & 59.63 & 3.15  \\
          &       & VPTQ   & 36.27 & 69.54 & 29.46 & 50.08 & 73.01 & 27.35 & 76.38 & 91.40 & 67.39 & 57.87 & 3.62  \\

\noalign{\vspace{0.2em}}\cdashline{2-14}\noalign{\vspace{0.2em}}

          & \multirow{7}[0]{*}{3.5} & RTN   & 39.59 & 72.22 & 33.55 & 56.31 & 74.29 & 31.4 & 78.01 & 92.3 & 69.92 & 60.84 & 3.31 \\
          &       & GPTQ   & 38.90 & 72.72 & 34.46 & 56.34 & 75.08 & 31.20 & 78.67 & 91.70 & 71.19 & 61.14 & 3.29 \\
          &       & AWQ    & 29.09 & 57.82 & 23.59 & 39.99 & 64.47 & 20.20 & 67.19 & 90.70 & 64.71 & 50.86 & 6.06 \\
          &       & QuaRot & 38.05 & 69.06 & 33.80 & 50.96 & 44.56 & 30.80 & 76.38 & 93.00 & 70.40 & 56.33 & 10.63   \\
          &       & KMeans & 41.29 & 74.07 & 34.82 & 54.76 & 70.66 & 31.60 & 78.23 & 94.30 & 70.63 & 61.15 & 3.84  \\
          &       & GPTVQ  & 36.65 & 70.87 & 31.91 & 52.43 & 75.04 & 30.00 & 77.31 & 93.40 & 70.24 & 59.76 & 3.34  \\
          &       & VPTQ   & 37.10 & 69.87 & 29.69 & 50.31 & 72.31 & 27.45 & 76.56 & 91.60 & 66.55 & 57.93 & 3.75 \\
          
\noalign{\vspace{0.2em}}\cdashline{2-14}\noalign{\vspace{0.2em}}
          & \multirow{1}[0]{*}{3.275}      & \cellcolor[rgb]{ .906,  .902,  .902}\textbf{Ours} 
          & \cellcolor[rgb]{ .906,  .902,  .902}41.80  
          & \cellcolor[rgb]{ .906,  .902,  .902}76.50 
          & \cellcolor[rgb]{ .906,  .902,  .902}34.09  
          & \cellcolor[rgb]{ .906,  .902,  .902}55.26  
          & \cellcolor[rgb]{ .906,  .902,  .902}78.17  
          & \cellcolor[rgb]{ .906,  .902,  .902}32.80 
          & \cellcolor[rgb]{ .906,  .902,  .902}78.61  
          & \cellcolor[rgb]{ .906,  .902,  .902}95.40
          & \cellcolor[rgb]{ .906,  .902,  .902}71.60  
          & \cellcolor[rgb]{ .906,  .902,  .902}62.69
          & \cellcolor[rgb]{ .906,  .902,  .902}2.89 \\
\noalign{\vspace{0.2em}}\hline\noalign{\vspace{0.2em}}
\hline\noalign{\vspace{0.2em}}
\end{tabular}}}
\end{table}

%% file: tabs/appendix_ablition_mul.tex
\begin{table*}[htbp]
\renewcommand\arraystretch{1.0}
\centering
\caption{\small Complete ablation study on the impact of codebook optimization for element-wise
multiplication.}
\vspace{-4em}
\label{tab:appendix_main_mul}
\setlength{\tabcolsep}{1mm}
{\resizebox{\textwidth}{!}{
\begin{tabular}{c|c|cccccccccc|c}
%& & & & & & & & & & & &\\
& & & & & & & & & & & &\\
& & & & & & & & & & & &\\
& & & & & & & & & & & &\\
\noalign{\vspace{0.2em}}\hline\noalign{\vspace{0.1em}}
\noalign{\vspace{0.1em}}\hline\noalign{\vspace{0.2em}}
\multirow{2}{*}{\textbf{Model}}  & \multirow{2}{*}{\textbf{Method}} & \textbf{ARC-c} & \textbf{ARC-e
} & \textbf{HQA.} & \textbf{HellaS.
} & \textbf{Lam.
} & \textbf{OBQA} & \textbf{PIQA
} & \textbf{SCIQ
} & \textbf{WinoG.
}  & \textbf{Avg.
}  & \textbf{Wiki2} \\ 
&  & ($\uparrow$) & ($\uparrow$) & ($\uparrow$) & ($\uparrow$) & ($\uparrow$) & ($\uparrow$) & ($\uparrow$) & ($\uparrow$) & ($\uparrow$) & ($\uparrow$) & ($\downarrow$) \\
\noalign{\vspace{0.2em}}\hline\noalign{\vspace{0.2em}}
\multirow{2}{*}{RWKV7-0.1B} 
         & w.   & 19.79 & 42.17 & 24.14 & 30.02 & 42.25 & 17.00 & 63.10 & 79.10 & 52.30 & 41.09 & 18.41  \\
         & wo.   & 20.15 & 44.45 & 24.47 & 30.32 & 38.09 & 17.80 & 64.08 & 73.95 & 52.91 & 40.69 & 24.71   \\
\noalign{\vspace{0.2em}}\hline\noalign{\vspace{0.2em}}
\multirow{2}{*}{RWKV7-0.5B} 
         & w.   & 22.61 & 53.11 & 25.41 & 34.45 & 53.39 & 19.60 & 66.53 & 84.20 & 54.80 & 46.01 & 9.39  \\
         & wo.   & 21.61 & 53.02 & 25.31 & 34.55 & 47.10 & 18.90 & 66.53 & 83.7 & 54.61 & 45.03 & 13.49   \\
\noalign{\vspace{0.2em}}\hline\noalign{\vspace{0.2em}}
\multirow{2}{*}{RWKV7-1.47B} 
         & w.    & 28.15 & 61.32 & 29.35 & 43.24 & 65.38 & 22.60 & 71.70 & 88.90 & 61.01 & 52.40 & 5.24  \\
         & wo.   & 29.15 & 61.90 & 29.99 & 42.92 & 59.31 & 24.10 & 70.13 & 88.52 & 64.14 & 52.24 & 6.54   \\
\noalign{\vspace{0.2em}}\hline\noalign{\vspace{0.2em}}
\multirow{2}{*}{RWKV6-1B} 
         & w.   & 27.90 & 60.35 & 28.08 & 42.51 & 63.87 & 23.00 & 71.38 & 87.9 & 60.22 & 51.69 & 5.29  \\
         & wo.   & 27.38 & 60.05 & 27.42 & 42.31 & 63.71 & 23.2 & 71.72 & 87.5 & 59.75 & 51.44 & 5.32    \\
\noalign{\vspace{0.2em}}\hline\noalign{\vspace{0.2em}}
\multirow{2}{*}{RWKV6-3B} 
         & w.   & 30.97 & 65.74 & 30.12 & 46.90 & 71.18 & 25.00 & 74.53 & 92.20 & 65.51 & 55.79 & 3.88  \\
         & wo.   & 30.97 & 68.00 & 29.72 & 46.84 & 70.46 & 25.20 & 74.21 & 90.80 & 63.14 & 55.48 & 3.97   \\
\noalign{\vspace{0.2em}}\hline\noalign{\vspace{0.2em}}
\multirow{2}{*}{RWKV6-7B} 
         & w.    & 37.62 & 72.93 & 33.63 & 52.03 & 75.70 & 30.40 & 76.6 & 95.1 & 67.71 & 60.19 & 3.23  \\
         & wo.   & 37.97 & 73.19 & 33.44 & 52.04 & 75.18 & 30.80 & 76.61 & 93.8 & 68.59 & 60.18 & 3.21   \\
\noalign{\vspace{0.2em}}\hline\noalign{\vspace{0.2em}}
\multirow{2}{*}{RWKV6-14B} 
         & w.  & 41.80 & 76.50 & 34.09 & 55.26 & 78.17 & 32.80 & 78.61 & 95.40 & 71.60 & 62.69 & 2.89  \\
         & wo.   & 40.53 & 73.82 & 34.13 & 56.19 & 77.72 & 32.00 & 78.44 & 94.00 & 71.50 & 62.03 & 2.89   \\

\noalign{\vspace{0.2em}}\hline\noalign{\vspace{0.2em}}
\hline%\noalign{\vspace{0.2em}}
% \vspace{-2mm}
\end{tabular}}}
\end{table*}

%% file: tabs/different_muf.tex
\begin{table}[H]
\renewcommand\arraystretch{0.8}
  \centering
  % \vspace{3mm}
  % 0-shot$^9$ includes ARC-easy, ARC-challenge, PIQA, and WinoGrande tasks, while 0-shot8adds BoolQ, SIQA, HellaSwag, LAMBADA(OpenAI) and OBQA tasks.
  \caption{Comparison of perplexity on LAMBADA and averaged accuracy on nine Zero-Shot tasks under different $\tau_c$ and $\tau_f$ configurations for the RWKV models.
  }
  \vspace{-2mm}
  \label{different_result}
  \setlength{\tabcolsep}{1mm}
  {
  \resizebox{0.7\textwidth}{!}{
    \begin{tabular}{c|c|cc:cc:cc}
    % \toprule
    & & & & & & &\\
\noalign{\vspace{0.1em}}\hline\noalign{\vspace{0.1em}}
\hline\noalign{\vspace{0.1em}}
\multirow{3}[2]{*}{\textbf{$\tau_c$}} & \multirow{3}[2]{*}{\ \  \textbf{$\tau_f$}\ \ } & \multicolumn{2}{c:}{\textbf{RWKV7-0.1B}} & \multicolumn{2}{c|}{\textbf{RWKV7-0.5B}} & \multicolumn{2}{c}{\textbf{RWKV7-1.47B}}
\\
\noalign{\vspace{0.1em}}\cdashline{3-8}\noalign{\vspace{0.1em}}
    % \midrule
     &  & 0-shot$^9$ & LambA.  & 0-shot$^9$ & LambA.  & 0-shot$^9$ & LambA.    \\
     &       & Avg.($\uparrow$) & ($\downarrow$)   & Avg.($\uparrow$) & ($\downarrow$)   & Avg.($\uparrow$) & ($\downarrow$)  \\
    % \midrule
    % \midrule
\noalign{\vspace{0.1em}}\hdashline\noalign{\vspace{0.1em}} 
\multirow{4}[2]{*} {1.00}
          & 20.00    & 40.27  & 23.65   & 43.72  & 13.98 & 52.09  & 5.52    \\
          & 25.00    & 40.27  & 23.65   & 43.72  & 13.98 & 52.09  & 5.52     \\
          & 30.00    & 40.27  & 23.65   & 43.62  & 14.06 & 52.09  & 5.52    \\       
          & 35.00    & 40.25  & 23.75   & 43.64  & 14.15 & 52.06  & 5.54     \\
          & 40.00    & 40.25  & 23.75   & 43.64  & 14.15 & 52.06  & 5.54     \\
\noalign{\vspace{0.1em}}\hdashline\noalign{\vspace{0.1em}} 
\multirow{4}[2]{*} {1.50}
          & 20.00    & 40.56  & 18.68   & 45.93  & 9.88  & 52.37  & 5.24     \\
          & 25.00    & 41.01  & 18.31   & 46.01  & 9.46  & 52.40  & 5.24      \\
          & 30.00    & 41.10  & 18.41   & 46.01  & 9.39  & 52.40  & 5.24    \\       
          & 35.00    & 40.86  & 18.71   & 46.01  & 9.45  & 52.40  & 5.24      \\
          & 40.00    & 40.78  & 18.41   & 45.88  & 9.67  & 52.34  & 5.24      \\
\noalign{\vspace{0.1em}}\hdashline\noalign{\vspace{0.1em}} 
\multirow{4}[2]{*} {2.00}
          & 20.00    & 39.86  & 28.12   & 45.93  & 9.88  & 52.34  & 6.12   \\
          & 25.00    & 39.86  & 28.12   & 45.93  & 9.88  & 52.34  & 6.12     \\
          & 30.00    & 39.65  & 31.34   & 45.87  & 10.07 & 52.26  & 6.32    \\      
          & 35.00    & 39.37  & 37.25   & 45.83  & 13.01 & 52.28  & 6.54     \\
          & 40.00    & 39.37  & 37.25   & 45.83  & 13.01 & 52.28  & 6.54     \\      
    % \bottomrule
    \noalign{\vspace{0.1em}}\hline\noalign{\vspace{0.1em}}
\hline\noalign{\vspace{0.1em}}
    \end{tabular}%
    }
    }
  \vspace{-2mm}
\end{table}%